\pdfoutput=1

\documentclass[11pt]{article}

\usepackage[]{EMNLP2022}

\usepackage{times}
\usepackage{latexsym}

\usepackage[T1]{fontenc}

\usepackage[utf8]{inputenc}

\usepackage{microtype}

\usepackage{inconsolata}

\usepackage{xspace}
\usepackage{booktabs}
\usepackage{graphicx}
\usepackage{enumitem}
\usepackage{xcolor,colortbl}
\usepackage{tikz-dependency}
\usepackage{url}
\usepackage{amssymb}
\usepackage{float}
\usepackage{todonotes}

\usepackage{caption}
\usepackage{subcaption}

\usepackage{tikz} %

\usepackage{rotating} %
\usepackage{graphics} %
\usepackage{multirow} %
\usepackage{bibentry} %

\nobibliography*

\newcommand{\dul}[1]{\underline{\smash{#1}}}

\setlist[description]{leftmargin=\parindent,labelindent=0pt}

\newcommand{\enquote}[1]{``#1''}

\newcommand{\sref}[1]{Sec.~\ref{sec:#1}}
\newcommand{\aref}[1]{Appendix~\ref{sec:#1}}
\newcommand{\srefplural}[1]{Sec.~\ref{sec:#1}}
\newcommand{\srefshort}[1]{\ref{sec:#1}}
\newcommand{\tref}[1]{Table~\ref{tab:#1}}
\newcommand{\fref}[1]{Figure~\ref{fig:#1}}

\definecolor{anne}{rgb}{0.635,0.998,0.722}
\definecolor{sophie}{rgb}{0.998,0.722,0.635}
\definecolor{stefan}{rgb}{0.635,0.722,0.998}
\definecolor{alex}{rgb}{0.722,0.635,0.998}

\definecolor{UniBlue}{rgb}{0.635,0.998,0.722}

\newcommand{\labelCapability}{\textit{\textbf{capability}}\xspace}
\newcommand{\labelDeontic}{\textit{\textbf{deontic}}\xspace}
\newcommand{\labelFeasibility}{\textit{\textbf{feasibility}}\xspace}
\newcommand{\labelInference}{\textit{\textbf{inference}}\xspace}
\newcommand{\labelRhetorical}{\textit{\textbf{rhetorical}}\xspace}
\newcommand{\labelUncertainty}{\textit{\textbf{speculation}}\xspace}
\newcommand{\labelOptions}{\textit{\textbf{options}}\xspace}

\newcommand{\labelCapabilityShort}{\textit{\textbf{cap.}}\xspace}
\newcommand{\labelDeonticShort}{\textit{\textbf{deon.}}\xspace}
\newcommand{\labelFeasibilityShort}{\textit{\textbf{feas.}}\xspace}
\newcommand{\labelInferenceShort}{\textit{\textbf{inf.}}\xspace}
\newcommand{\labelRhetoricalShort}{\textit{\textbf{rhet.}}\xspace}
\newcommand{\labelUncertaintyShort}{\textit{\textbf{spec.}}\xspace}
\newcommand{\labelOptionsShort}{\textit{\textbf{opt.}}\xspace}

\newcommand{\corpusName}{\textsc{MiST}\xspace}
\newcommand{\corpusSizeModalInstancesTotal}{3737\xspace}

\newcommand{\fscore}{F\textsubscript{1}\xspace}
\newcommand{\macrofscore}{mF\textsubscript{1}\xspace}

\newcounter{example}
\newenvironment{example}[1][]{\refstepcounter{example}\par
	\textbf{Example~\theexample. #1} \rmfamily}{}

\usepackage{array}
\newcolumntype{L}[1]{>{\raggedright\let\newline\\\arraybackslash\hspace{0pt}}m{#1}}
\newcolumntype{C}[1]{>{\centering\let\newline\\\arraybackslash\hspace{0pt}}m{#1}}
\newcolumntype{R}[1]{>{\raggedleft\let\newline\\\arraybackslash\hspace{0pt}}m{#1}}

\title{\corpusName: a Large-Scale Annotated Resource and Neural Models\\ for Functions of Modal Verbs in English Scientific Text}

\author{Sophie Henning$^{1,2}$ \hspace{5mm} Nicole Macher$^{1}$ \hspace{5mm} Stefan Grünewald$^{1,3}$ \hspace{5mm} Annemarie Friedrich$^{1}$ \\
 $^1$Bosch Center for Artificial Intelligence, Renningen, Germany \\
 $^2$Center for Information and Language Processing, LMU Munich, Germany \\
 $^3$Institut für Maschinelle Sprachverarbeitung, University of Stuttgart, Germany\\
  \texttt{sophieelisabeth.henning@de.bosch.com} \hspace{2mm}
  \texttt{macher.nicole@gmail.com} \\
  \texttt{stefan.gruenewald|annemarie.friedrich@de.bosch.com} \\
}

\begin{document}
\maketitle
\begin{abstract}
Modal verbs (e.g., \textit{can}, \textit{should} or \textit{must}) occur highly frequently in scientific articles.
Decoding their function is not straightforward: they are often used for hedging, but they may also denote abilities and restrictions.
Understanding their meaning is important for various NLP tasks such as writing assistance or accurate information extraction from scientific text.
	
To foster research on the usage of modals in this genre, we introduce the \corpusName (\textbf{M}odals \textbf{I}n \textbf{S}cientific \textbf{T}ext) dataset, which contains \corpusSizeModalInstancesTotal modal instances in five scientific domains annotated for their semantic, pragmatic, or rhetorical function.
We systematically evaluate a set of competitive neural architectures on \corpusName.
Transfer experiments reveal that leveraging non-scientific data is of limited benefit for modeling the distinctions in \corpusName.
Our corpus analysis provides evidence that scientific communities differ in their usage of modal verbs, yet, classifiers trained on scientific data generalize to some extent to unseen scientific domains.
\end{abstract}
\section{Introduction}
\label{sec:intro}

Each year, an estimate of 1.5 million scientific articles are published \citep{wosp-2020-international}; hence, the construction of knowledge graphs (KGs) from scholarly texts for aggregating and navigating research findings is an active research area \citep{sdp-2020-scholarly,wosp-2020-international, ws-2019-extracting,ws-2019-bionlp,bionlp-2020-sigbiomed}.
Professional academic writing makes ample use of \textit{hedges}, linguistic devices indicating uncertainty,
because scientific propositions are usually considered as valid only until they are overwritten by newer findings \citep{hyland1998hedging}.
Distinguishing valid solutions to problems from unverified and/or potential solutions is a crucial step in information extraction (IE) from scientific text \citep{heffernan2018identifying} as KGs %
should at least mark untested hypotheses as such (see \fref{teaser-img}).
Yet, with the notable exception of BioScope \citep{szarvas-etal-2008-bioscope}, prior work in this area is limited.

\begin{figure}
	\footnotesize
	
	\colorbox{gray!50}{Internal Property} \labelCapability\\	
\textit{This sensor \dul{\textbf{can}} both respond to pressure and tension.}\\
\begin{tikzpicture}
\node[rounded rectangle, draw, thick] (a) at (0,0) {sensor\_X};
\node[rounded rectangle, draw, thick] (b) at (5, 0) {respond\_to\_pressure};
\node[rounded rectangle, draw=gray, thick, dotted, text=darkgray] (c) at (6.1, -.5) {true};
\node[] (d) at (2.2, 0) {\textit{hasCapabilityTo}};
\node[text=darkgray] (e) at (3, -.5) {\textit{hasFactualityRating}};
\draw [->, thick] (a) -- (d)-- (b);
\draw [->, thick, draw=darkgray, dotted] (e) -- (c);
\draw [-, thick, draw=darkgray, dotted] (1.4, -.1) -- (1.4, -0.5);
\draw [-, thick, draw=darkgray, dotted] (1.4, -.5) -- (1.7, -0.5);
\end{tikzpicture}
	
\colorbox{gray!50}{External action} \labelFeasibility\\
\textit{Graphene polymer composites films \dul{\textbf{can}} be used as protective elements in electronics.}\\
\vspace*{-6mm}

\begin{tikzpicture}
\node[rounded rectangle, draw, thick, text width=2.1cm] (a) at (0,0) {graphene\_poly- mer\_composites};
\node[rounded rectangle, draw, thick] (f) at (3.5, 0.6) {electronic\_device};
\node[rounded rectangle, draw, thick] (b) at (5, 0) {protective\_element};
\node[rounded rectangle, draw=gray, thick, dotted, text=darkgray] (c) at (5.5, -.55) {possible};
\node[] (d) at (2.5, 0) {\textit{composedOf}};
\node[] (g) at (5.5, 0.6) {\textit{partOf}};
\node[text=darkgray] (e) at (3, -.5) {\textit{hasFactualityRating}};
\draw [->, thick] (b) -- (d)-- (a);
\draw [->, thick, draw=darkgray, dotted] (e) -- (c);
\draw [-, thick, draw=darkgray, dotted] (2.4, -.1) -- (2.4, -0.4);
\draw [-, thick, draw] (5.5, .25) -- (5.5, .5);
\draw [->, thick, draw] (g) -- (f);
\end{tikzpicture}
\vspace*{-4mm}

	\colorbox{gray!50}{Hypothesis} \labelUncertainty\\
\textit{The inferior results \dul{\textbf{may}} be related to catalyst particles.}\\ 
	\begin{tikzpicture}
	\node[rounded rectangle, draw, thick] (a) at (0,0) {catalyst\_particles};
	\node[rounded rectangle, draw, thick] (b) at (5, 0) {results\_paper\_X};
	\node[rounded rectangle, draw=gray, thick, dotted, text=darkgray] (c) at (5.5, -.55) {speculation};
	\node[] (d) at (2.55, 0) {\textit{cause}};
	\node[text=darkgray] (e) at (2.5, -.55) {\textit{hasFactualityRating}};
	\draw [->, thick] (a) -- (d)-- (b);
	\draw [->, thick, draw=darkgray, dotted] (e) -- (c);
	\draw [-, thick, draw=darkgray, dotted] (2.5, -.1) -- (2.5, -0.55);
	\end{tikzpicture}

	\caption{\textbf{Modal verbs} perform various \textbf{functions in scientific text} affecting KG representations.}
	\label{fig:teaser-img}
\end{figure}
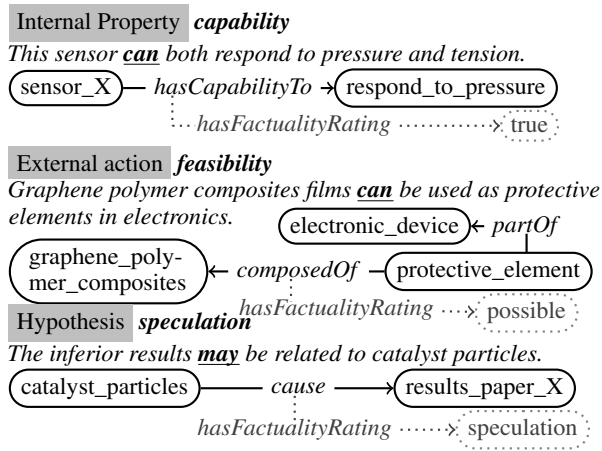

In this paper, we focus on \textbf{modal verbs}, a frequently used device for signaling hedging in academic discourse \cite{hanania1985verb,getkham2011hedging}.
Other functions of modals include indicating abilities or restrictions.
Their meaning depends on the sociopragmatic context \citep{yamazaki2001pragmatic}, i.e., here on the conventions of the community of a particular academic field.
Successful academic writing requires correct community-specific use.
As shown in \fref{teaser-img}, understanding the different notions has relevance to KG population \citep[e.g.,][]{luan-etal-2018-multi,friedrich-etal-2020-sofc}.
Computational modeling of the functions of modal verbs also has applications in language learning and writing assistance software \citep{roemer2004corpus}.

Prior work in computational linguistics targeting modal verbs \citep{ruppenhofer-rehbein-2012-yes,rubinstein-etal-2013-toward,pyatkin-etal-2021-possible,marasovic-etal-2016-modal} has primarily worked with data from the news domain.
The annotation schemes of these datasets largely follow distinctions established in the linguistic literature \citep[see, e.g.,][]{kratzer1981notional,palmer2001mood,fintel2006modality,portner2009modality}, differentiating between the following coarse-grained modal senses: (a) \textit{epistemic} expresses judgments about the factual status of a proposition, (b) \textit{deontic} relates to permission, obligation, and requirements, and (c) \textit{dynamic} refers to internal abilities or conditions.
Our work differs from all of these approaches (a) in that we are the first to address the domain of scientific writing, and (b) in that we do not primarily study modal \textit{senses}, but instead focus on the \textit{pragmatic function} of modal verbs, i.e., our aim is to capture an author's \textit{reason} for using a particular modal verb in a context.

With this paper, we release \corpusName (\textbf{M}odals \textbf{I}n \textbf{S}cientific \textbf{T}ext), a manually annotated \textbf{corpus} for investigating the usage of modal verbs in scientific text.
Our \textbf{multi-label} annotation scheme for \textit{modal functions} covers semantic, pragmatic, and rhetorical reasons for an author's use of a modal, with a focus on sub-distinctions that are crucial from an IE viewpoint.
\corpusName consists of \corpusSizeModalInstancesTotal annotated modal verb instances selected from texts of five scientific disciplines (henceforth \textit{domains}), which is larger than all existing comparable datasets (see \tref{datasets-overview}).
Our \textbf{corpus analysis} reveals differences in modal use between scientific domains, and between academic and non-academic use.
We perform an inter-annotator agreement study and ensure high data quality via adjudication.

Based on \corpusName, as well as related corpora, we conduct an extensive \textbf{computational study} on automatically classifying functions of modals, comparing CNN-based \citep{marasovic-frank-2016-multilingual} and BERT-based models \citep[similar to][]{pyatkin-etal-2021-possible}. %
In contrast to prior modeling work, we circumvent modifying the transformer's input by selecting the modal's contextualized output embedding and/or the CLS embedding as input to the classifier.
We find that in most cases, a model using both embeddings works best.

To sum up, our paper lays the groundwork for both corpus-linguistic and computational work on modeling functions of modal verbs in scientific text. %
Our contributions are as follows.
\begin{description}
	\setlength{\itemsep}{1pt}
	\setlength{\parskip}{0pt}
	\setlength{\parsep}{0pt}
	\item[$\bullet$] Our new large-scale dataset annotated with functions of modals in scientific text is publicly available.\footnote{\href{https://github.com/boschresearch/mist_emnlp_findings2022}{github.com/boschresearch/mist\_emnlp\_findings2022}}
	\item[$\bullet$] We conduct an in-depth corpus study detailing the corpus construction process, agreement, and corpus statistics, as well as a comparison with existing schemes (\sref{corpus-analysis}).
	\item[$\bullet$] Our computational experiments provide a systematic comparison of neural models for modal classification on scientific text (\srefplural{modeling} and \srefshort{experiments}).
	We find that a combination of the CLS embedding and the embedding of the modal verb itself works best. %
	\item[$\bullet$] We show that models trained on out-of-genre data do not work well on scientific text, while classifiers trained on annotated scientific text perform well on unseen scientific domains. In sum, these experimental findings underline the value of our new dataset.
\end{description}

\section{Related work}
\label{sec:relwork}

Our work relates to several areas, which we survey in this section.

\noindent\textbf{Annotated corpora.}
Prior annotation studies on modal verb senses carried out by expert annotators are of limited scale (see \tref{datasets-overview}). %
\citet[][henceforth RR12, \textbf{Modalia} dataset]{ruppenhofer-rehbein-2012-yes} annotate \textit{senses} of modal verbs in the MPQA Opinion corpus, which consists of news texts. %
Their linguistically motivated label set includes \textit{dynamic}, \textit{epistemic}, and \textit{deontic} (see \sref{intro}), as well as \textit{optative} for wishes, \textit{concessive} if a state of affairs is taken as a given, and \textit{conditional} for \textit{if}-clauses and inversion constructions.
On the same texts, \citet[][henceforth Rubin13]{rubinstein-etal-2013-toward} annotate modal expressions including nouns (e.g., \enquote{hope}), adjectives, adverbs, and verbs of propositional attitude (e.g., \enquote{believe}).
Their annotation scheme is similar to RR12 with minor modifications. %
\citet[][henceforth Pyatkin21]{pyatkin-etal-2021-possible} use this dataset with six renamed categories and refer to it as the Georgetown Gradable Modal Expressions (\textbf{GME}) corpus.
\citet{marasovic-etal-2016-modal} annotate a 3-way distinction of modal senses (\textit{dynamic}, \textit{epistemic}, and \textit{deontic}) on \textbf{MASC} \citep{ide-etal-2008-masc}, covering several domains. %
They also introduce the Modalia version \textbf{Modalia\textsubscript{M}} using this 3-way scheme, mapping \textit{conditional} and \textit{concessive} to \textit{epistemic} and \textit{optative} to \textit{deontic}.
Finally, their \textbf{EPOS} dataset consists of 7693 sentences for which the same 3-way annotation has been derived via cross-linguistic projection from Europarl \citep{koehn-2005-europarl} and OpenSubtitles \citep{tiedemann-2012-parallel}.
\citet{king-morante-2020-must} annotate modal verbs in the vaccination debate domain (\textbf{VCM}).

\begin{table}[t]
	\centering
	\footnotesize
	\setlength{\tabcolsep}{3pt}
	\begin{tabular}{lrrll}
		\toprule
		\textbf{Dataset} & \textbf{\# inst.} & \textbf{\# cat.} & \textbf{genre} & \textbf{lang.} \\ %
		\midrule
		Modalia &  1158 & 6 & news & EN \\ %
		Rubin13/GME\textsuperscript{*} & 1912 & 6 & news & EN\\
		MASC & 1962 & 3 & multi-genre & EN\\ %
		VCM & 450 & 6 & vaccination debate & EN\\
		EP & 888 & 13 & multi-genre & PT \\
		CuiChi13 & 263 & 6 & news & CN\\
		\midrule
		\corpusName (ours) & \corpusSizeModalInstancesTotal & 7 & scientific papers & EN\\ %
		\bottomrule
	\end{tabular}
	\caption{\textbf{Datasets manually annotated with \textbf{modal verb} categories.} \textsuperscript{*}Rubin13: 6 base categories + 3 supertypes + 1 multi-label combination;  GME is the same dataset as Rubin13 using the 6 base categories (renamed) and 2 supertypes. For Rubin13/GME, EP, and CuiChi13, we count only modal verb instances. %
	}
	\label{tab:datasets-overview}
\end{table}

\pagebreak
Several annotated datasets target modal expressions in a variety of domains, e.g., focusing on \textit{could} \citep{moon-etal-2016-selective} in English GigaWord \citep{parker2009english}, or negotiation dialogues \citep{lapina-petukhova-2017-classification}.
We are also aware of a cluster of works on annotating and tagging Portuguese data (\textbf{EP}) using multi-genre data and RR12-style annotation schemes \citep[e.g.,][]{mendes2016modality,avila-mello-2013-challenges,quaresma2014tagging}.
\citet{cui-chi-2013-annotating} conduct a small annotation study on Chinese modals with Rubin13-style labels (\textbf{CuiChi13}).
\citet{yamazaki2001pragmatic} performs a corpus study on how American English native speakers interpret modal verbs in the chemistry domain.

\noindent\textbf{Modeling.}
Early approaches to modal sense classification leverage a lexicon \citep{baker-etal-2010-modality}, or make use of \enquote{traditional} features (such as n-grams or part-of-speech tags) and maximum entropy classifiers \citep{ruppenhofer-rehbein-2012-yes,zhou-etal-2015-semantically} or SVMs \citep{quaresma2014automatic,quaresma2014tagging}.
\citet{li2019modal} create context vectors for modals by computing weighted sums of the non-contextualized word embeddings of selected context words.
\citet[][henceforth MF16]{marasovic-frank-2016-multilingual} generate a sentence embedding using a CNN, hence classifying \textit{sentences} instead of \textit{modal instances}.
Our models are most similar to those of Pyatkin21, who encode input sentences using RoBERTa \citep{liu2019roberta}, with the CLS embedding as input for a linear classifier.
Their model variants differ in the input: the \textit{Context} model marks the modal trigger with special tags (\texttt{Sue <target>can</target> swim}); %
the \textit{Trigger+Head} model encodes only the trigger and its dependency head without further context.
 
\noindent\textbf{Further related work.}
Other related work includes research on speculation in biomedical data \cite{szarvas-etal-2008-bioscope,kim-etal-2011-overview-genia} and on event factuality \citep[e.g.,][]{sauri2009factbank,stanovsky-etal-2017-integrating,rudinger-etal-2018-neural-models,pouran-ben-veyseh-etal-2019-graph}.
\citet{bijl-de-vroe-etal-2021-modality} integrate a lexicon-based method for modality detection in event extraction; using this tagger, \citet{guillou2021blindness} find that entailment graph construction does not profit from tagging for modality.
\citet{vigus-etal-2019-dependency} propose to annotate modal structures as dependencies.
Rhetorical analysis of scientific text is often based on Argumentative Zoning \citep{teufel1999argumentative}.
\citet[][]{lauscher-etal-2018-argument,lauscher-etal-2018-investigating} provide a dataset and neural methods for extracting and classifying claims from scientific text.
\citet{luan-etal-2018-multi}, \citet{jiang2019role}, and \citet{friedrich-etal-2020-sofc} present data-driven work on scientific IE.
\citet{heffernan2021problem} uses modality as a feature to recognize problem-solving utterances in scientific text.

\section{\corpusName Corpus}
\label{sec:corpus-analysis}

In this section, we describe our new dataset, including its annotation scheme and detailed corpus and inter-annotator agreement statistics.
We annotate instances of \textit{can, could, may, might, must}, and \textit{should} %
in research papers from five scientific fields: computational linguistics (CL), materials science (MS), agriculture (AGR), earth science (ES), and computer science (CS).
Modal usage is influenced by sociopragmatic context \citep{yamazaki2001pragmatic} and, as a form of hedging, needs to be understood in its social, cultural and institutional context \citep{hyland1998hedging}, here the \textit{global} scientific community.
Hence, we do not restrict document selection to native English authors.

\begin{table}[t]
	\centering
	\footnotesize
	\setlength{\tabcolsep}{2pt}
	\begin{tabular}{l|rrrrr|r}
\toprule
& CL &   MS &  AGR &   CS &   ES &  Total \\
\midrule
\multicolumn{2}{l}{\textit{Complete corpus}} \\
sent with modals &   925 &  718 &  497 &  746 &  584 &   3470 \\
annotated modals &  1011 &  757 &  543 &  806 &  620 &   3737 \\
\midrule
\multicolumn{3}{l}{\textit{Full-text subset}}\\
documents             &   30 &   16 &   10 &    7 &   10 &     73 \\
sents. with modals       &  693 &  462 &  195 &  445 &  258 &   2053 \\
- in \% of sents.*     &  9.3 & 11.6 &  9.0 & 14.2 & 11.0 &   10.8 \\
sents. with $\geq$ 2 modals &   61 &   29 &   24 &   48 &   19 &    181 \\
avg. \#tokens/sent.       & 27.7 & 26.4 & 32.1 & 27.4 & 31.9 &   28.4 \\
annotated modals                 &  760 &  492 &  223 &  497 &  279 &   2251 \\
\bottomrule
\end{tabular}

	\caption{\textbf{\corpusName corpus statistics}. *in \% of total sentences of the documents. In the \textit{full-text subset}, all modals within the documents have been annotated. The \textit{complete corpus} contains the full-text subset and additionally sampled individual sentences with annotations.}
	\label{tab:corpus_stats_base}
\end{table}

\subsection{Document and Sentence Selection}
\label{sec:dataset-appendix}
We select modal verb occurrences as follows.
In our \textit{full-text subset} of 73 documents, the CL papers are taken from the ACL Anthology,\footnote{\href{https://aclanthology.org}{aclanthology.org}} spanning the years 2013-2015. 
Data from the other domains stems from the OA-STM corpus,\footnote{\href{https://elsevierlabs.github.io/OA-STM-Corpus}{elsevierlabs.github.io/OA-STM-Corpus}} with the exception of five open-access documents for MS.

Because some modal-domain combinations are rare, we additionally sample sentences from 348 documents with Creative Commons licenses such that we have at least 100 instances for each modal-domain pair.
For CS, we sample papers tagged with \textit{cs:CV} and published in 2018 from ArXiv.\footnote{\href{https://kaggle.com/Cornell-University/arxiv}{kaggle.com/Cornell-University/arxiv}}
Additional MS papers published between 2015 and 2021 were retrieved via PubMedCentral.\footnote{\href{https://ncbi.nlm.nih.gov/pmc}{ncbi.nlm.nih.gov/pmc}}
For ES and AGR, we use the DOAJ API\footnote{\href{https://doaj.org/api/v2/docs}{doaj.org/api/v2/docs}} to retrieve documents matching the topics of the full-text subset.
For AGR, we add articles from the Journal of Agricultural Science published 2017-2021.\footnote{\href{https://cambridge.org/core/journals/journal-of-agricultural-science}{cambridge.org/core/journals/journal-of-agricultural-science}}
In total, we obtain a large-scale dataset of \corpusSizeModalInstancesTotal annotated instances (see \tref{corpus_stats_base}, \textit{complete corpus}).

\begin{table*}[t]
	\centering
	\footnotesize
	\setlength{\tabcolsep}{2pt}
	\begin{tabular}{llll}
		\toprule
		Example & Ours & RR12 & Rubin13 / Pyatkin21\\
		\midrule
		Several supercapacitors \dul{\textbf{can}} be integrated and connected in series. & \labelFeasibility & \textit{dynamic} & \textit{Circum.} / \textit{State of the World}\\
		The device \dul{\textbf{can}} light up a red light-emitting diode and works well. &\labelCapability &  \textit{dynamic} & \textit{Ability} / \textit{State of the Agent} \\
		The overlap in the ranges [...] indicates that the sample \dul{\textbf{must}} be & \labelInference & \textit{epistemic} & \textit{Epistemic} / \textit{State of Knowledge} \\
		\hspace*{5mm}older than 50.70 Ma. \\
		The real shielding \dul{\textbf{can}} of course be different. 
		 & \labelOptions &  \textit{deontic} & \textit{Circum.} / \textit{State of the World}\\
		DA3 \dul{\textbf{may}} therefore indicate a continuation of high nutrient surface & \labelUncertainty  &  \textit{epistemic} & \textit{Epistemic} / \textit{State of Knowledge} \\
		\hspace*{5mm}water with an elevated freshwater input.
		& \\
		Energy storage devices \dul{\textbf{should}} be able to endure high-level strains. & \labelDeontic  & \textit{deontic} & \textit{Bouletic} / \textit{Desires and Wishes} \\
		A GCR proton [...] \dul{\textbf{must}} have at least 150 MeV to reach the station. & \labelDeontic  & \textit{deontic} & \textit{Teleological} / \textit{Plans and Goals} \\
		You must leave the lab tidy. & \labelDeontic & \textit{deontic} & \textit{Deontic} / \textit{Rules and Norms} \\
		It \dul{\textbf{can}} be seen in Figure 1 that... & \labelRhetorical  & \textit{dynamic} & \textit{Ability} / \textit{State of the Agent} \\
		\midrule
		For instance, despite graphene, the band gaps of silicone \dul{\textbf{can}} be\\
		\hspace*{5mm}opened and tuned when exposed to an external electric field. & \labelFeasibilityShort, \labelCapabilityShort & \textit{dynamic} & \textit{Circum.} / \textit{State of the World} \\
		These results suggest that epeiric seas [...] \dul{\textbf{may}} have played an & \\
		\hspace*{5mm}important role in the driving mechanism for OAE 2. & \labelInferenceShort, \labelUncertaintyShort & \textit{epistemic} & \textit{Epistemic} / \textit{State of Knowledge} \\
		\midrule
		\textit{Long \dul{\textbf{may}} she live!} & \labelDeontic & \textit{optative} & \textit{Bouletic} / \textit{Desires and Wishes} \\
		\bottomrule
	\end{tabular}
	\caption{\textbf{\corpusName \textbf{annotation scheme}} in comparison to those of RR12 and Rubin13/Pyatkin21.} %
	\label{tab:scheme_comparison}
\end{table*}

\subsection{Annotation Scheme}
Our annotation scheme comprises seven labels for functions of modals (see \tref{scheme_comparison}).
\tref{label_applicability} shows which labels apply for each modal, for more details see \aref{details-annotation-scheme}.
The labels apply if the author uses the modal verb:\vspace*{-2mm}

\begin{description}
\setlength{\itemsep}{1pt}
\setlength{\parskip}{0pt}
\setlength{\parsep}{0pt}
	\item[\labelFeasibility:] to indicate that it is possible for an external actor, e.g., a human, to do or achieve something; 	
	\item[\labelCapability:] to convey that something has a certain intrinsic property, ability, or capacity;	
	\item [\labelInference:] to state that they inferred something based on some given information; %
	\item [\labelUncertainty:] to indicate speculations;
	\item [\labelOptions:] to indicate potential options; %
	\item [\labelDeontic:] to express a desire, or a requirement, or an obligation;
	\item [\labelRhetorical:] for conventionalized, fixed expressions. %
\end{description}\vspace*{-2mm}

Our inventory intends to capture the most frequent and relevant functions of modal verbs in scientific discourse.
Table \ref{tab:scheme_comparison} classifies a set of utterances according to our, RR12's and Rubin13's schemes.\footnote{According to our interpretation of their guidelines.
To facilitate comparison with Pyatkin21, we also added their mapping to the Rubin13 scheme to the table.}
A detailed description of the commonalities and differences %
is provided in \aref{scheme-comparison}.
During annotation scheme design, we started out with their categories, but then tailored our scheme to the scientific domain, adding some pragmatic distinctions that are relevant in scientific writing.
Annotators have access to the full documents.
For labels involving inference, uncertainty or speculation,
annotators are instructed to only refer to the text and not to make use of their own knowledge of whether something is the case.

\begin{figure}[t]
	\centering
	\includegraphics[width=0.48\textwidth]{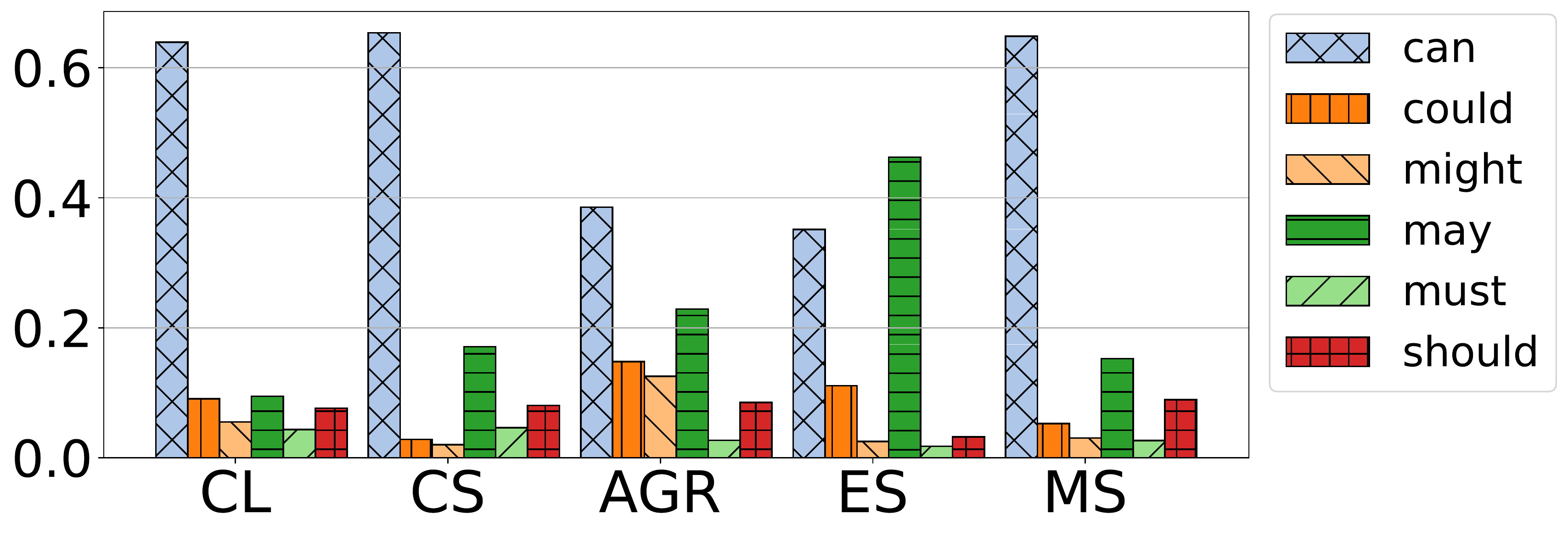}
	\caption{\textbf{\corpusName:} Distribution of modals by domain, computed over full-text annotation subset.} %
	\label{fig:modal-dist-by-domain}
\end{figure}

\begin{figure*}[t]
	\centering
	\begin{subfigure}{.02\textwidth}
		\footnotesize
		\rotatebox{90}{\hspace*{10mm}Percentage}
	\end{subfigure}
	\begin{subfigure}{.13\textwidth}
		\centering
		\includegraphics[height=24mm]{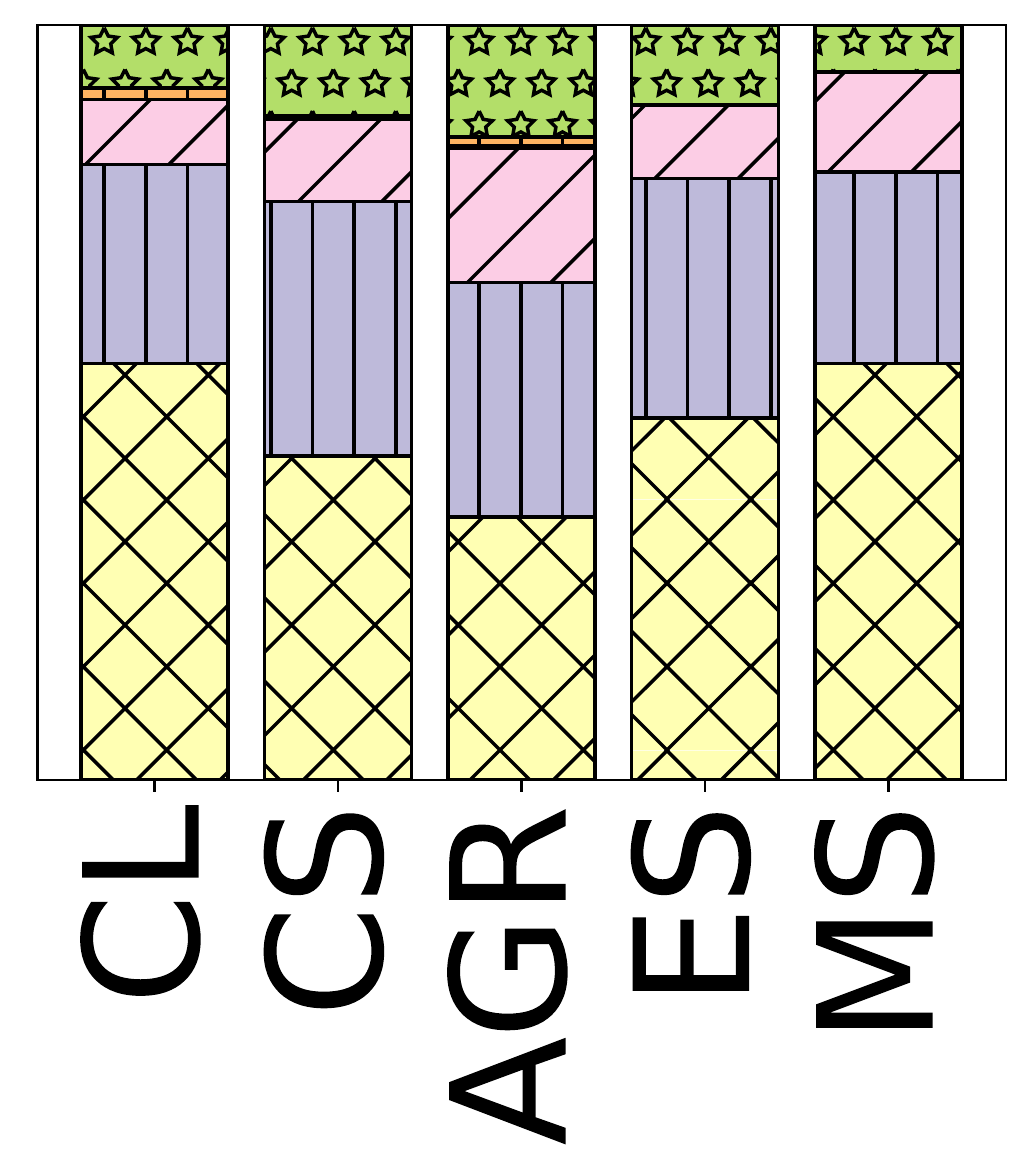}
		\caption{can}
	\end{subfigure}
	\begin{subfigure}{.13\textwidth}
		\centering
		\includegraphics[height=24mm]{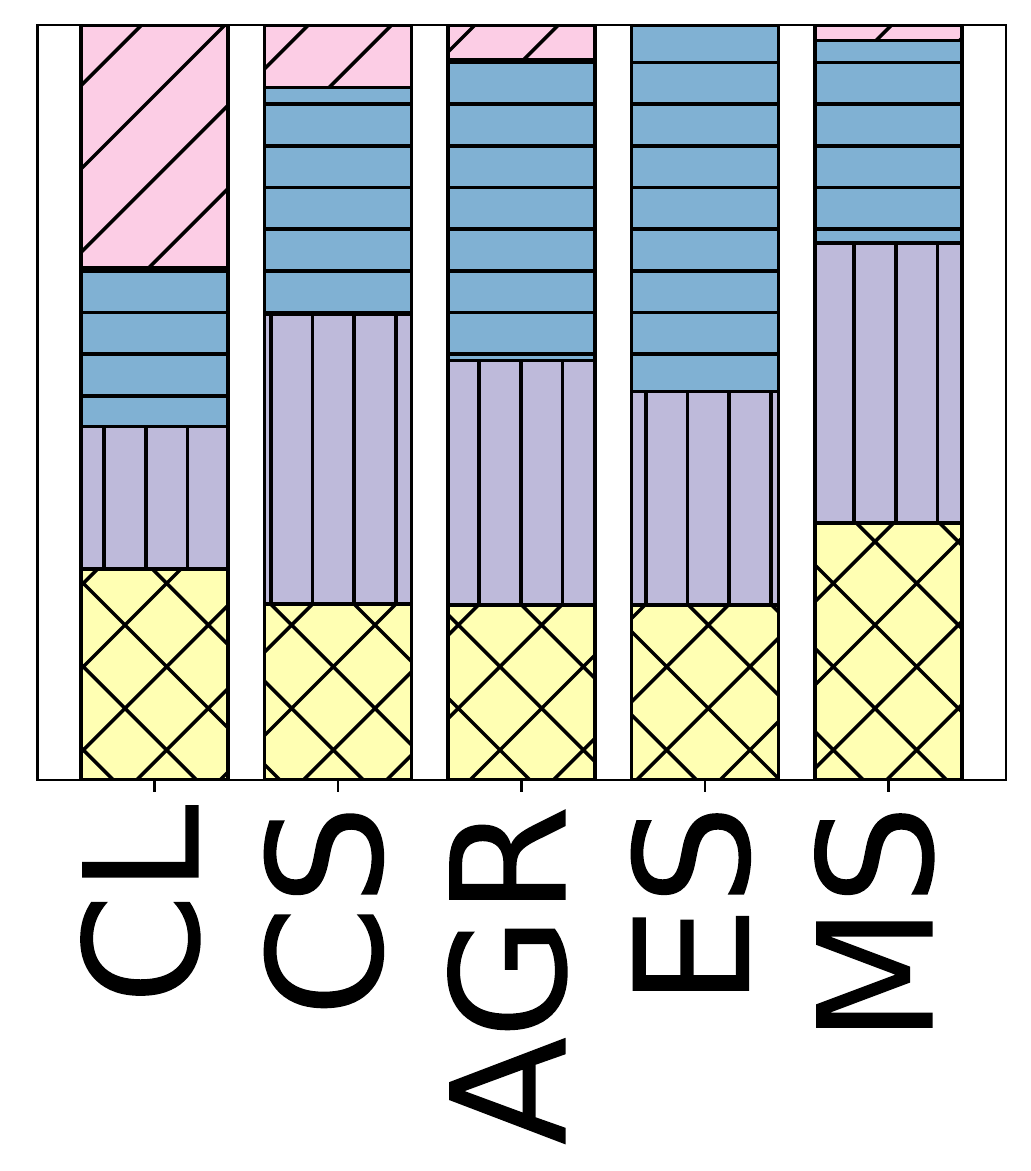}  
		\caption{could}
	\end{subfigure}
	\begin{subfigure}{.13\textwidth}
		\centering
		\includegraphics[height=24mm]{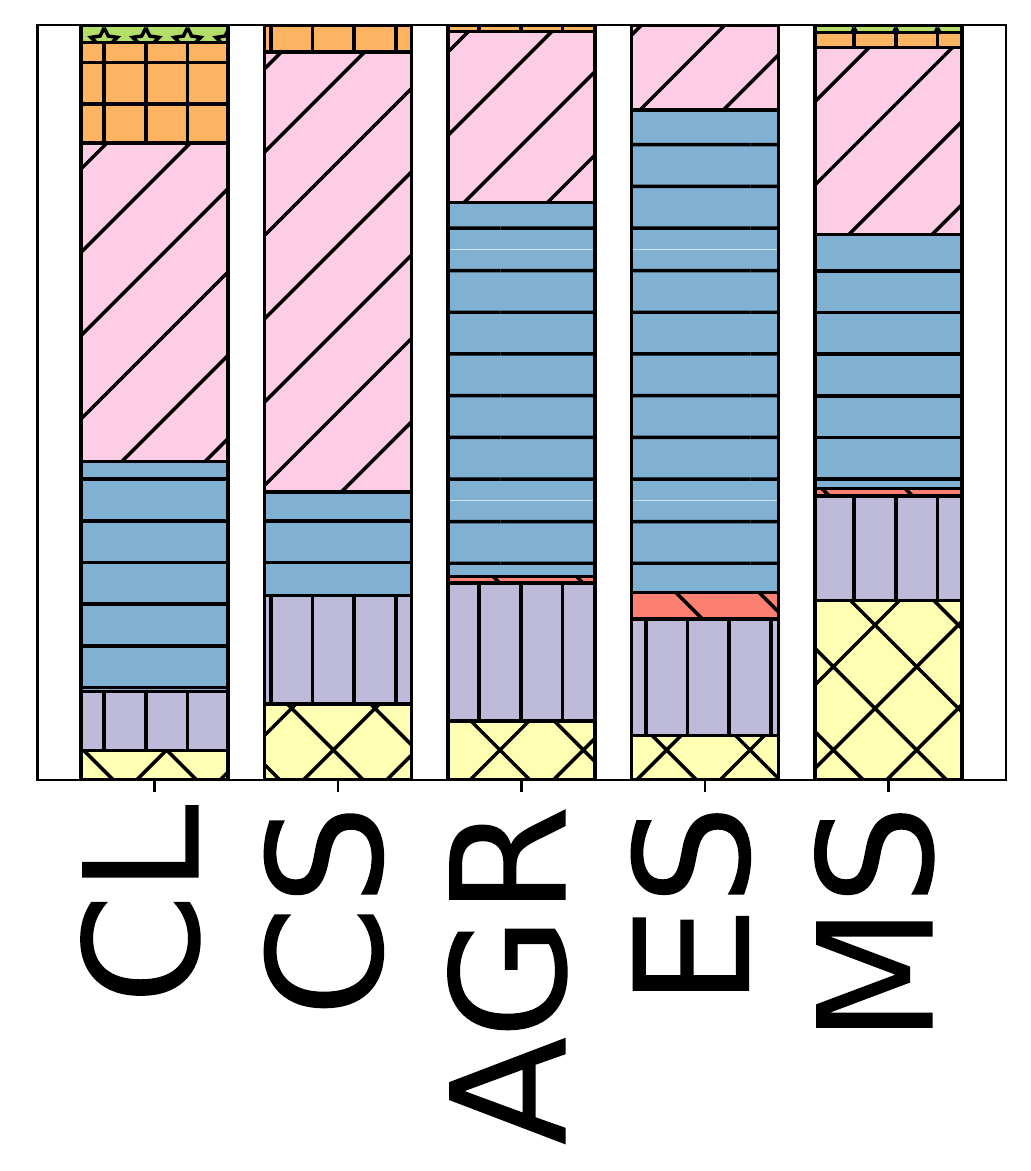}  
		\caption{may}
	\end{subfigure}
	\begin{subfigure}{.13\textwidth}
		\centering
		\includegraphics[height=24mm]{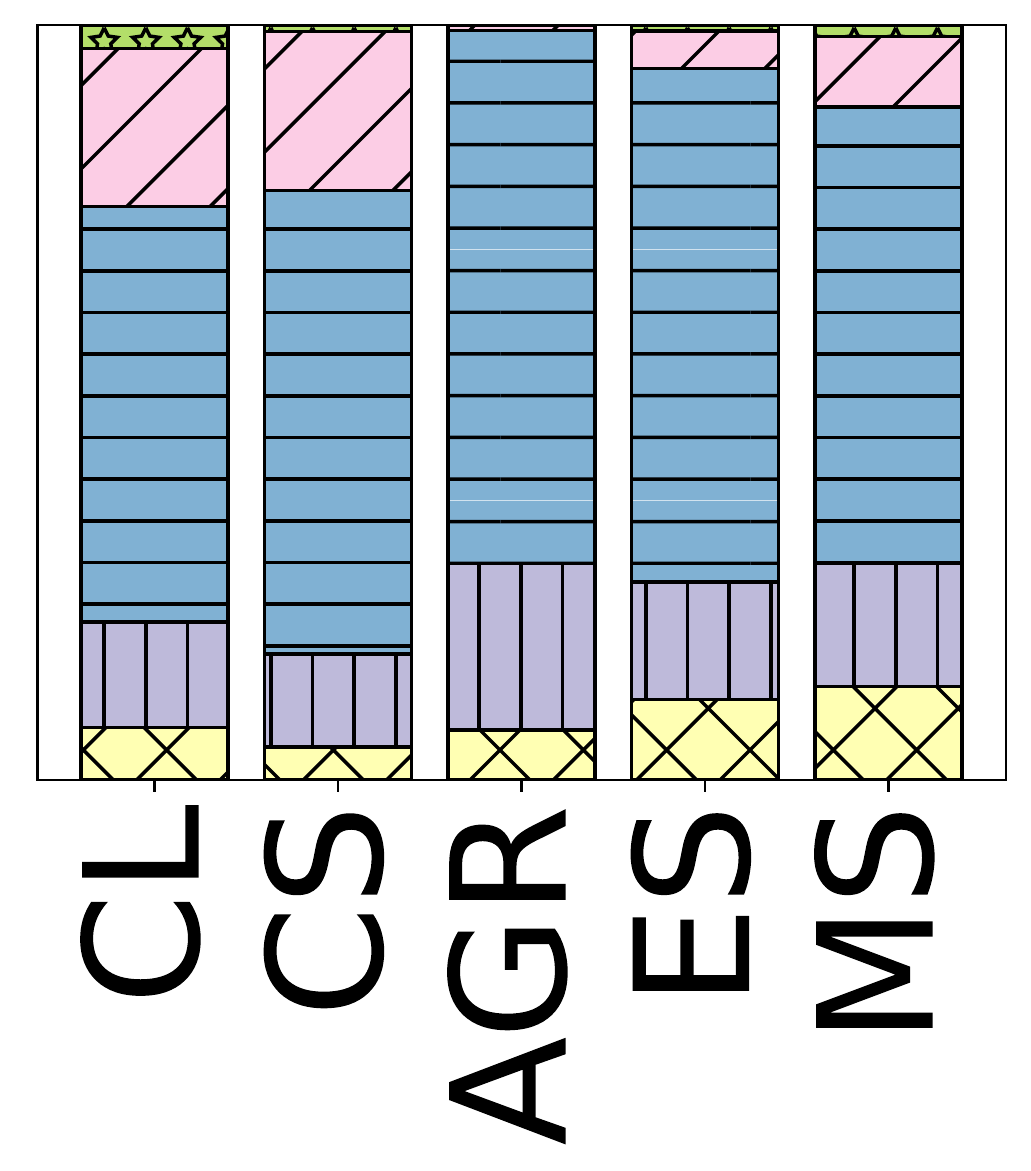}  
		\caption{might}
	\end{subfigure}
	\begin{subfigure}{.13\textwidth}
		\centering
		\includegraphics[height=24mm]{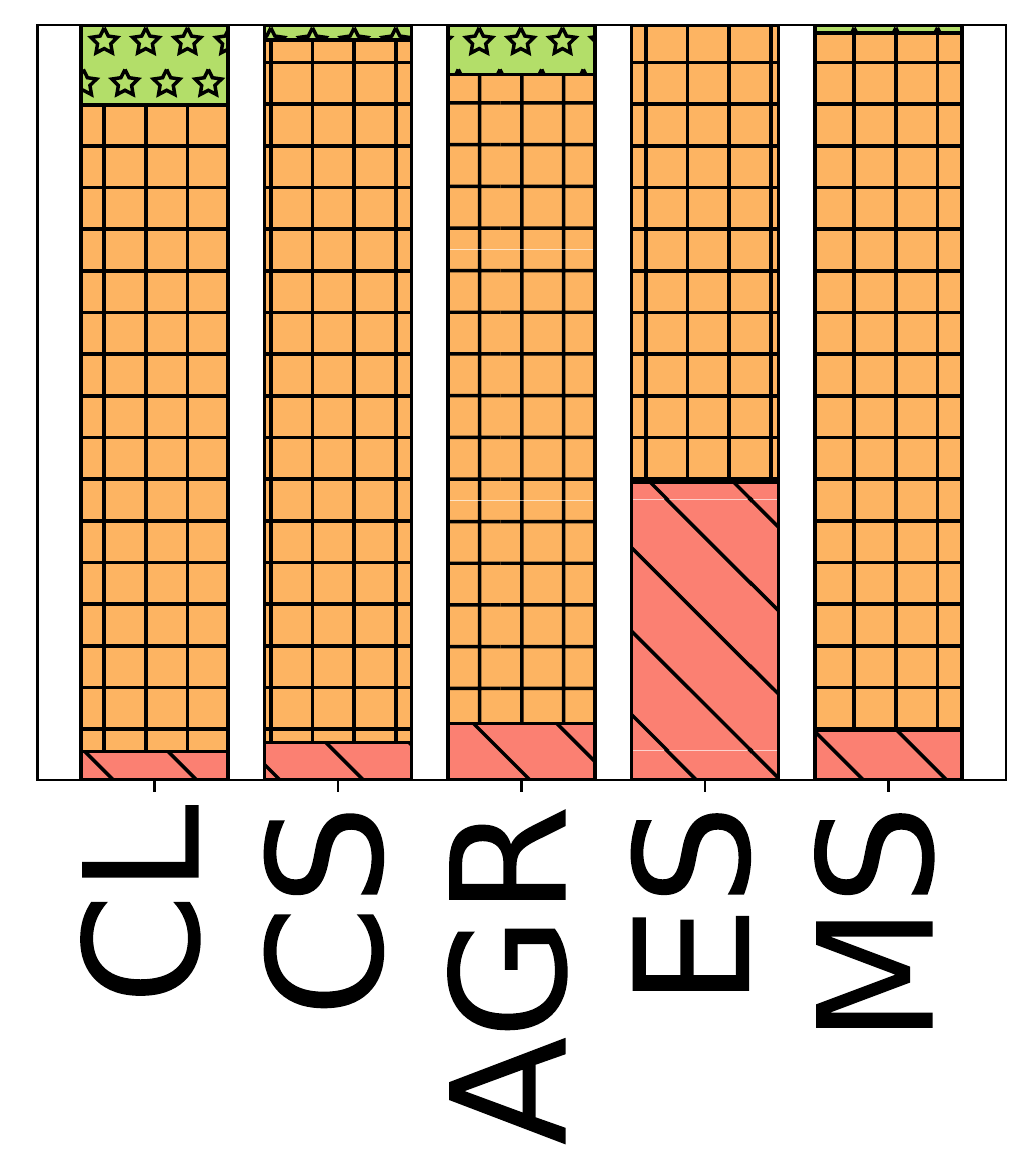}  
		\caption{must}
	\end{subfigure}
	\begin{subfigure}{.28\textwidth}
		\includegraphics[height=24mm]{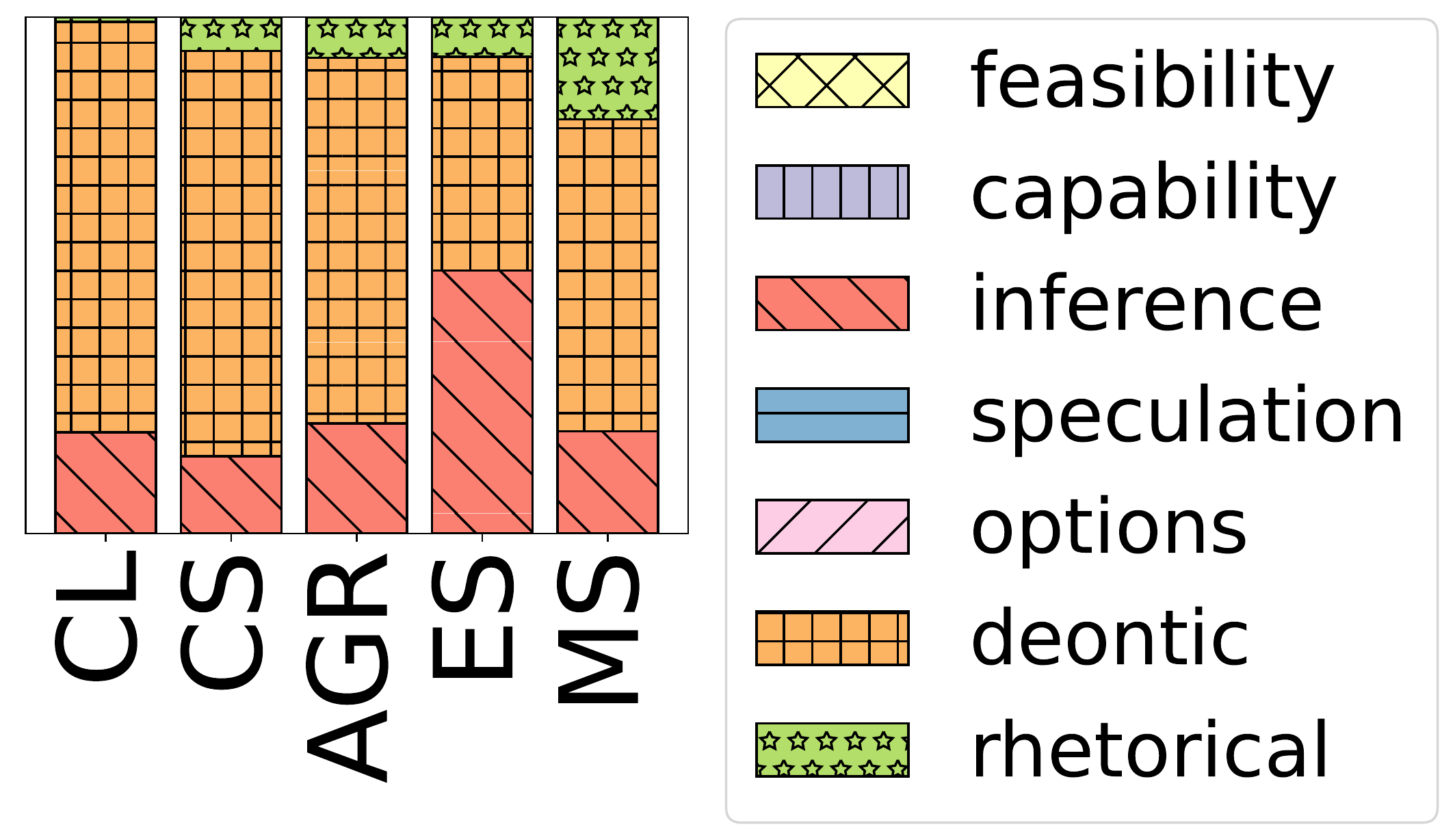}  
		\caption{should\hspace*{2cm}}
	\end{subfigure}
	\caption{\textbf{\corpusName: Label distributions} by modal verb and scientific domain (adjudicated complete corpus).} %
	\label{fig:label_dist_genre_modal}
\end{figure*}

\subsection{Annotation Process}
Our annotation scheme takes a multi-label approach in which all applicable features may be selected.
For each instance of the \textit{full-text subset}, we collect the annotations of three annotators (two for MS) using the web-based annotation systems Swan \citep{guehring2016swan} and INCEpTION \citep{klie-etal-2018-inception}.
We ensure consistency across sub-corpora by means of an adjudication step (for all instances) performed by one author of this paper, who then also labeled the additionally sampled instances.
Our total group of annotators consists of one undergraduate as well as three graduate students of CL, one undergraduate student of CS, one graduate student of MS and one physicist holding a PhD degree.
While not all annotators are native speakers of English, they are either domain experts or have a strong linguistic background.

\subsection{Corpus Analysis}
\label{sec:corpus_analysis}

\textbf{Modal distributions.}
We first analyze the usage of the different modals per domain.
As shown in \tref{corpus_stats_base}, in the full-text subset, the ratio of sentences including modal verbs ranges from 9.0\% (AGR) to 14.2\% (CS).
In \fref{modal-dist-by-domain}, we plot the distributions of modals by domain.
Except in the case of ES, \textit{can} is the most frequently used modal by a large margin.
In AGR and ES, \textit{may} is also used frequently.
Overall, the distributions of CL, CS and MS are somewhat similar, while AGR and ES exhibit different modal usage patterns.
The distributions differ from modal usage in other genres (details for MASC and Modalia see \aref{masc-modalia-stats}), e.g., the percentage of \textit{can} is much higher in \corpusName.

\noindent\textbf{Label distributions.}
Next, we drill down on the functions of the modals by domain.
If an instance has more than one label, both labels are counted.
The label distributions differ strongly by modal (see \fref{label_dist_genre_modal} and \tref{label_applicability}), but at times also visibly between domains.
Previous corpus-linguistic studies \citep{takimoto2015corpus,hardjanto2016hedging} observe more hedging in humanities and social sciences text compared to the natural sciences.
ES, which deals with earth's present features and its past evolution, has notably more \labelInference usages of \textit{must} and \textit{should}. %
In MS, many cases of \textit{could} are classified as \labelFeasibility, as it is common to report experiments in the past tense in this domain.
Also, in MS \textit{may} is sometimes used interchangeably with \textit{can} as in \enquote{stress–strain data \dul{may} be obtained for ductile materials.}
The larger amount of \labelRhetorical instances in MS is due to cases such as \enquote{We \dul{should} note that.}

Comparing the label distributions of \corpusName and those of MASC and Modalia\textsubscript{M}, we also find notable differences (details in \aref{masc-modalia-stats}).
For example, \textit{may} is used mostly in \textit{epistemic} senses.
Our annotations reveal that in AGR and ES, these are mostly \labelUncertainty; CL and CS texts use this modal to indicate (mostly algorithmic) \labelOptions. 
Finally, the use of \textit{should} seems most community-specific: while it is used predominantly in a \textit{deontic} way in MASC and Modalia\textsubscript{M}, usage in \corpusName varies by domain.
Overall, these observations support the hypothesis that modal usage depends on the sociopragmatic context, and demonstrate the value of genre-specific data such as \corpusName. %

\begin{table}[t]
	\centering
	\footnotesize
	\setlength{\tabcolsep}{3pt}
	\renewcommand{\arraystretch}{0.8}
	\begin{tabular}{rrrrrrr}
		\toprule
		& can & could & may & might & must & should\\
		\midrule
		\labelFeasibility &  823 &    161 &   62 &     52 &     0 &       0 \\
		\labelCapability  &  476 &    188 &   91 &    102 &     0 &       0 \\
		\labelInference   &    0 &      0 &    *8 &      0 &    63 &     127 \\
		\labelUncertainty &    0 &    206 &  257 &    398 &     0 &       0 \\
		\labelOptions     &  183 &     64 &  205 &     70 &     0 &       0 \\
		\labelDeontic     &   *13 &      0 &   25 &      0 &   444 &     330 \\
		\labelRhetorical  &  157 &      0 &    *4 &     *8 &    24 &      41 \\
		\bottomrule
	\end{tabular}
	\caption{\textbf{\corpusName: Label counts}, all domains, adjudicated complete corpus. *Omitted from experiments.
	}
	\label{tab:label_applicability}
\end{table}

\noindent\textbf{Label co-occurrence.}
In the full-text subset and in the complete corpus 24.5\% and 22.3\% of instances carry more than one label, respectively.
\fref{label_cooccurrence} shows the total number of label co-occurrences in the adjudicated gold standard.
Overall, \labelUncertainty co-occurs most with other labels, indicating that the author likely had two reasons for using the modal, for example indicating a \labelCapability, but marking at the same time that it is unclear whether it actually holds
(\enquote{The urban ecosystems \dul{could} account for a significant portion of terrestrial carbon (C) storage (...).}). %
Often, both a \labelFeasibility and a \labelCapability reading are possible (see lower part of \tref{scheme_comparison}), as in \enquote{The above construction \dul{can} be further simplified.}, where simplifiability is an intrinsic property of the construction, but the simplification needs an external actor.

\begin{figure}[t] %
	\centering
	\footnotesize
	\setlength\tabcolsep{4pt}
	\begin{tabular}{rC{8mm}C{8mm}C{8mm}C{10mm}C{8mm}C{8mm}}
		\labelCapabilityShort & \cellcolor{white!26!gray}153 &    \\
		\labelInferenceShort	 & 0 & \cellcolor{white!99!gray}1 &  \\
		\labelUncertaintyShort & \cellcolor{white!65!gray}73 &  \cellcolor{white!0!gray}208 &  \cellcolor{white!95!gray}8 & \\
		\labelOptionsShort & \cellcolor{white!68!gray}67 & \cellcolor{white!44!gray}117 & 0 & \cellcolor{white!99!gray}1\\
		\labelDeonticShort &	 \cellcolor{white!99!gray}2 &   \cellcolor{white!99!gray}2 &    \cellcolor{white!93!gray}15 & 0 & \cellcolor{white!93!gray}15  \\
		\labelRhetoricalShort &	 \cellcolor{white!38!gray}128 &    \cellcolor{white!99!gray}3 &   \cellcolor{white!99!gray}2 &  \cellcolor{white!97!gray}5 &  6 & \cellcolor{white!76!gray}50 \\
		& \labelFeasibilityShort & \labelCapabilityShort & \labelInferenceShort & \labelUncertaintyShort & \labelOptionsShort & \labelDeonticShort \\
	\end{tabular}
	\caption{\textbf{\corpusName: Label co-occurrence counts}, all domains, adjudicated complete corpus.}
	\label{fig:label_cooccurrence}
\end{figure}

\subsection{Inter-Annotator Agreement}
\label{sec:iaa-f1}
Computing agreement for our dataset is not straightforward for two reasons.
First, we are dealing with a multi-label scenario, for which standard agreement coefficients cannot easily be applied.
Second, for some modal-domain combinations, we only have limited data.
Averaging across modal verbs is not meaningful: due to the notably different label distributions, good agreement could only mean that annotators distinguish modal verbs well \citep{artstein2009semiformal,artstein2017inter}.

Following the idea of Krippendorff's diagnostics \citep{krippendorff1980content}, we evaluate (on the full-text subset) for each modal-label combination how often annotators agree on whether the label applies or not.
For each pair of annotators, we compute $\kappa$ \citep{cohen1960kappa} for this binary decision for each label, mapping all respective other labels to \textsc{Other}.
In \fref{iaa-plot}, we report the average of these $\kappa$-scores over the pairs of annotators for each valid modal-label combination.
For some combinations, high agreement is reached.
For infrequent labels or modals, agreement is less satisfying.
Many \enquote{disagreements} occur in cases where in fact several readings are possible.

\begin{figure}[t]
	\centering
	\includegraphics[width=0.48\textwidth]{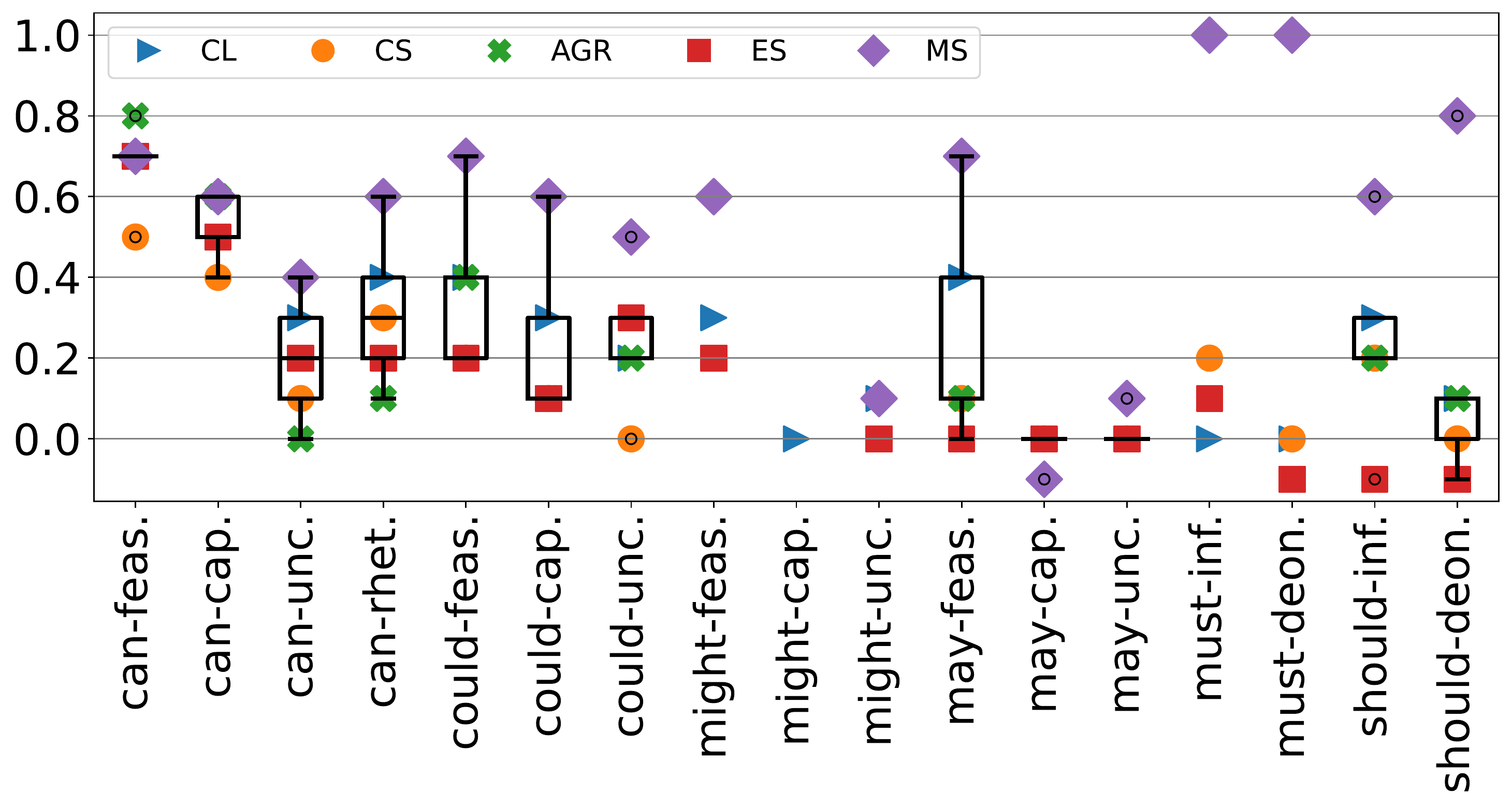}
	\caption{\textbf{\corpusName: Inter-annotator agreement} in terms of avg. $\kappa$ for labels that have been assigned to at least 6 instances of respective modal (by any annotator), on full-text subset.
	} %
	\label{fig:iaa-plot}
\end{figure}

Qualitative analysis revealed that some annotators over- or under-used some labels, especially \textbf{\textit{uncertainty}}, which in the initial round of annotation described here was defined to include both \labelOptions and \labelUncertainty.
We hence decided to \textbf{ensure high quality} of our corpus through an \textbf{adjudication step}.
In 62.2\% of instances, the adjudicator's labels exactly match the majority vote across annotators; in 90.5\%, they overlap with the majority vote labels.
We further introduced the label \labelOptions, and two adjudicators re-labeled all instances initially labeled with \labelUncertainty.
Out of these, both labeled 166 instances, reaching F1-agreements of 72.7/81.3/83.5/86.9 for \labelCapability, \labelFeasibility, \labelOptions and \labelUncertainty, respectively.
In the remainder of this paper, we perform experiments based on the adjudicators' labels.

\section{Computational Modeling}
\label{sec:modeling}
We now describe our neural models for classifying functions of modal verbs.
We assume that targets have been pre-defined, e.g., using a part-of-speech tagger.
Our models are based on a pre-trained transformer that provides embeddings for sentences and contextualized token embeddings.
We fine-tune SciBERT \citep[SB,][]{beltagy-etal-2019-scibert}, which has the same architecture as BERT \citep{devlin-etal-2019-bert}, but has been trained on large volumes of scientific text. %
On top, we use \textbf{multiple classification heads}, i.e., one per modal, as the label distributions vary substantially by modal.
The largest version of our models is trained jointly on multiple datasets and therefore has the aforementioned output heads for each dataset (see \fref{model_architecture}).
The output dimension of these heads varies according to the labelset size of the respective dataset.

We test the following model variants:

\noindent
\textbf{SB\tiny{CLS}.}
We feed the CLS embedding of an input sentence into a linear layer with softmax (for single-label classification) or sigmoid (for multi-label classification) activation.
This model uses the same decision basis for all modal verbs within a single sentence.

\noindent
\textbf{SB\tiny{modal}.}
We select the embedding of the word-piece token corresponding to the modal to be classified (\textit{modal embedding}),\footnote{SciBERT and BERT both tokenize all modals in \corpusName into a single word piece. (Otherwise, one could use the embedding of the modal's first word-piece token.)} and feed this embedding into the linear layer as above.
We expect this model to be able to distinguish different modal verbs in the same sentence. %
The model primarily reflects local context, but to some extent also dependency context \citep{tenney-etal-2019-bert}.

\noindent
\textbf{SB\tiny{CLS,modal}.}
We concatenate the CLS embedding with the modal embedding before feeding it into the linear layer.
This model should distinguish modal verbs in the same sentence, at the same time leveraging the CLS embedding that intends to cover the entire sentence.%

\begin{figure}[t]
	\footnotesize
	\centering
	\begin{tikzpicture}[node distance=0.4cm and 0.05cm, scale=1, every node/.style={scale=0.8}, minimum size=0.5cm]	
	\node[rectangle, draw, align=center] (i) at (0,0) {concat};
	\node[align=center] (i2) [right=of i] {Instance embedding};
	\node[rectangle, draw] (s1) [below left=of i] {<CLS>};
	\node[rectangle, draw] (s2) [right=of s1] {BERT};
	\node[rectangle, draw, font=\bfseries] (s3) [right=of s2] {can};
	\node[rectangle, draw] (s4) [right=of s3] {embed};
	\node[rectangle, draw] (s5) [right=of s4] {tokens};
	\node[text width=2cm, align=center] (s6) [left=of s1] {SciBERT embeddings};
	
	\node[rectangle, draw, text width=1cm] (acl) [above left=of i] {\corpusName can};
	\node[text width=2cm, align=center]  (h) [left=of acl] {Classification heads};
	\node[scale=0.3] (p1) [right=of acl] {\Huge{...}};
	\node[rectangle, draw, text width=1cm] (ms) [right=of p1] {\corpusName should};
	\node[rectangle, draw, text width=1cm] (masc1) [right=of ms] {MASC can};
	\node[scale=0.3] (p2) [right=of masc1] {\Huge{...}};
	\node[rectangle, draw, text width=1cm] (masc2) [right=of p2] {MASC should};
	
	\node[text width=1.25cm, font=\itshape] (o) [above=of acl] {\textbf{\textit{\normalsize capability}}};
	
	\draw[->]
	(s1) edge (i) (s3) edge (i) (i) edge (acl) (i) edge (ms) (i) edge (masc1) (i) edge ++(25mm, 5.8mm) (masc2);
	\draw[->] (acl) edge ++(0, 8mm) (o);
	\end{tikzpicture}

	\caption{\textbf{Model architecture.} SB{\tiny CLS,modal} model.} %
	\label{fig:model_architecture}
\end{figure}
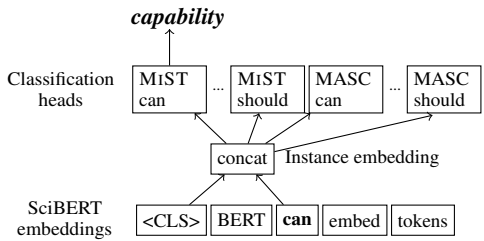

\section{Experiments}
\label{sec:experiments}
In this section, we report our experimental results. %

\subsection{Evaluation Metrics}
\label{sec:eval_metrics}
As majority classifiers are known to provide a strong baseline for modal sense classification (see Rubin13, MF16), we report \fscore scores in order to evaluate how well a classifier performs across labels.
We compute macro-average \fscore (\textbf{\macrofscore}) as the average of the per-label \fscore scores for the set of labels with which the modal is labeled at least once in the entire corpus and which are not omitted from the experiments due to extreme sparsity (see \tref{label_applicability}).
We also report accuracy; we compute it globally across samples and labels, i.e., we simply count for each label how often the classifier (in)correctly did (not) assign it.
For hyperparameter tuning and early stopping, we use the macro-average of weighted \fscore scores for each modal-domain combination.
These weighted \fscore scores are computed by weighting per-label \fscore scores by the label's support in the validation set.
For computing all metrics, we use TorchMetrics.\footnote{\href{https://github.com/PyTorchLightning/metrics}{github.com/PyTorchLightning/metrics}}
\nocite{torchmetrics} %

\subsection{Baselines}
We report results for the following baselines:
\textbf{Maj} always predicts the label most frequent in training.
We also re-implement MF16's \textbf{CNN} with 300-dimensional GloVe embeddings \citep{pennington-etal-2014-glove} and filter region sizes of 3, 4, and 5 with 100 filters each. 
 	Replicating MF16's Table 4 (with their hyperparameters and training a separate model for each modal), we find that our CNN implementation is comparable to theirs, with 77.6 \% accuracy on all verbs compared to MF16's 76.5\%. %
	On \corpusName, we use only one model with per-modal heads. %
	\textbf{SB\tiny{CLS-mark}} is our re-implementation of Pyatkin21's \textit{Context} model (their most accurate model), but using SciBERT and per-modal heads. %
	We also investigate whether the genre-specific pre-training is beneficial, replacing SciBERT with BERT (\textbf{BERT\tiny{CLS,modal}}), and how the model size affects performance, comparing to \textbf{BERT-large\tiny{CLS,modal}} (to date, there is no SciBERT-large). %

\subsection{Experimental Settings}
\label{sec:exp-setting}

We randomly split \corpusName into a training and a test set of complete documents, aiming at covering approximately 25\% of each domain's modal instances in the test set, with real test set sizes ranging from 22.8\% to 27\%.
In our \textit{CV training setting}, we split the training set into 5 folds of complete documents, and train 5 models on 4 folds each, using the respective fifth fold for model selection.
We train for at most 100 epochs, performing early stopping with a patience of 10 epochs.
We then run each of these five models on the unseen test set, reporting average scores along with standard deviations.
Hyperparameters are reported in \aref{hyperparameters}.

\subsection{Experimental Results on \corpusName}
\label{sec:mainresults}

Here, we evaluate the neural architectures described above on \corpusName, and investigate performance in the absence of in-domain training data.

\begin{table}[t]
	\centering
	\footnotesize
	\setlength{\tabcolsep}{4pt}
	\renewcommand{\arraystretch}{1.05}
	\begin{tabular}{l|cccccc}
		\toprule
		& \textbf{can} & \textbf{could} & \textbf{may} & \textbf{might} & \textbf{must} & \textbf{should}\\
		\midrule
		\#inst. train & 987 & 343 & 369 & 366 & 397 & 342 \\
		\#inst. test & 340 & 105 & 141 & 117 & 105 & 119 \\
		\midrule
		Maj & 18.9 & 15.5 & 12.8 & 23.0 & 30.2 & 28.9 \\[-0.15cm]
		& \tiny{$\pm0.0$} & \tiny{$\pm0.4$} & \tiny{$\pm0.0$} & \tiny{$\pm0.0$} & \tiny{$\pm0.0$} & \tiny{$\pm0.0$}\\
		CNN & 58.8 & 55.2 & 40.2 & 37.8 & 41.1 & 64.2\\[-0.15cm]
		& \tiny{$\pm5.5$} & \tiny{$\pm7.1$} & \tiny{$\pm5.4$} & \tiny{$\pm4.4$} & \tiny{$\pm4.2$} & \tiny{$\pm10.0$}\\
		SB\tiny{CLS} & 74.8 & 71.9 & 50.1 & 64.1 & 78.2 & 82.5\\[-0.15cm]
		& \tiny{$\pm2.1$} & \tiny{$\pm4.0$} & \tiny{$\pm3.1$} & \tiny{$\pm4.3$} & \tiny{$\pm4.3$} & \tiny{$\pm2.7$}\\
		SB\tiny{CLS-mark} & 76.6 & 63.7 & 49.1 & 61.5 & 73.7 & 85.5\\[-0.15cm]
		& \tiny{$\pm1.7$} & \tiny{$\pm1.4$} & \tiny{$\pm1.8$} & \tiny{$\pm1.5$} & \tiny{$\pm4.0$} & \tiny{$\pm2.3$}\\
		SB\tiny{modal} & 76.7 & 71.3 & \textbf{50.2} & \textbf{65.3} & 76.9 & 84.5\\[-0.15cm]
		& \tiny{$\pm2.5$} & \tiny{$\pm1.8$} & \tiny{$\pm4.3$} & \tiny{$\pm4.7$} & \tiny{$\pm2.7$} & \tiny{$\pm1.4$}\\
		SB\tiny{CLS, modal} & 77.4 & \textbf{73.7} & 47.2 & 64.5 & \textbf{78.4} & \textbf{85.7}\\[-0.15cm]
		& \tiny{$\pm1.0$} & \tiny{$\pm3.8$} & \tiny{$\pm1.1$} & \tiny{$\pm2.7$} & \tiny{$\pm1.1$} & \tiny{$\pm0.5$}\\
		BERT & 74.9 & 73.3 & 47.9 & 64.1 & 73.8 & 85.4\\[-0.15cm]
		\hspace*{3mm}\tiny{CLS,modal} & \tiny{$\pm2.2$} & \tiny{$\pm2.0$} & \tiny{$\pm1.6$} & \tiny{$\pm1.4$} & \tiny{$\pm3.4$} & \tiny{$\pm1.0$}\\
		BERT-large & \textbf{77.7} & 68.9 & 46.2 & 61.4 & 76.0 & 84.9\\[-0.15cm]
		\hspace*{3mm}\tiny{CLS,modal}& \tiny{$\pm1.3$} & \tiny{$\pm2.5$} & \tiny{$\pm3.0$} & \tiny{$\pm3.2$} & \tiny{$\pm2.0$} & \tiny{$\pm1.1$}\\
		\bottomrule
	\end{tabular}%
	\caption{\textbf{Macro \fscore (\macrofscore) on test set of \corpusName.}
		\#inst. train refers to the entire training set.
	} 
	\label{tab:evaluation_modals_scientific_f1}
\end{table}

\paragraph{Comparing Model Architectures.}
\tref{evaluation_modals_scientific_f1} reports the \macrofscore scores of the various neural models on \corpusName. 
The magnitude of these scores differs by modal verb. %
The CNN learns more than Maj., but is always outperformed by the SciBERT-based models.
SB{\tiny modal} is better than SB{\tiny CLS} on \textit{can}, \textit{might}, and \textit{should}, but worse on \textit{must}, where using an additional sentence-wide embedding is beneficial. %
For most of the verbs, SB{\tiny CLS,modal} is the best SciBERT-based model, but SB{\tiny modal} is better on \textit{may} and \textit{might}.
In general, SB{\tiny CLS, modal} tends to have smaller standard deviations across CV training configurations than the other SciBERT-based models.
On \textit{could} and \textit{must}, SB{\tiny CLS,modal} is better than SB{\tiny CLS-mark}, suggesting that directly using the modal's embedding instead of modifying the input is more effective. %

On most verbs, SciBERT and BERT perform comparably, but the domain specificity of SciBERT leads to clear improvements on \textit{can} and \textit{must}.
Interestingly, increasing the model size for BERT is beneficial on the very same verbs; at the same time, however, it hurts performance on the other verbs, with an especially sharp loss on \textit{could}.

With the exception of \textit{can}, SB{\tiny CLS,modal} is also the most accurate model (scores in \aref{further-results}).
For this model, during development, we experimented with using only one classifier head for all modals (not reported in tables).
Compared to per-modal heads, we observed either no difference or slightly worse (by around 1 point \macrofscore on average) performance for all modals except \textit{must}, where \macrofscore increased by around 15 points.
These gains were due to similar \labelRhetorical instances, e.g., \enquote{We \dul{must} note that...} and \enquote{We \dul{should} note that...}. %

\begin{table}[tb]
	\centering
	\footnotesize
	\setlength\tabcolsep{7pt}
	\setlength{\tabcolsep}{3pt}
	\begin{tabular}{c|ccccc}
		\toprule
		\textbf{Macro \fscore} & \textbf{CL} & \textbf{CS} & \textbf{Agr} & \textbf{ES} & \textbf{MS} \\
		\midrule
		+ & \textbf{57.0} & \textbf{53.2} & \textbf{61.7} & \textbf{58.7} & \textbf{59.7} \\[-0.15cm]
        & \tiny{$\pm4.6$} & \tiny{$\pm5.6$} & \tiny{$\pm8.1$} & \tiny{$\pm3.1$} & \tiny{$\pm1.1$}\\
		- & 54.2 & 50.2 & 60.4 & 54.8 & 58.8 \\[-0.15cm]
        & \tiny{$\pm1.7$} & \tiny{$\pm7.6$} & \tiny{$\pm7.0$} & \tiny{$\pm1.8$} & \tiny{$\pm5.1$}\\
		\midrule
		\textbf{Accuracy} & \textbf{CL} & \textbf{CS} & \textbf{Agr} & \textbf{ES} & \textbf{MS} \\
		\midrule
		+ & \textbf{92.3} & \textbf{92.7} &  \textbf{93.5} & \textbf{93.5} & \textbf{93.4} \\[-0.15cm]
        & \tiny{$\pm1.0$} & \tiny{$\pm1.4$} & \tiny{$\pm1.3$} & \tiny{$\pm0.8$} & \tiny{$\pm0.5$}\\
		- & 91.5 & 91.9 & \textbf{93.5} & 92.5 & 92.5 \\[-0.15cm]
        & \tiny{$\pm1.3$} & \tiny{$\pm0.9$} & \tiny{$\pm1.6$} & \tiny{$\pm0.9$} & \tiny{$\pm1.4$}\\
		\bottomrule
	\end{tabular}%
	\caption{Results for 6-fold CV on \corpusName by domain
		when \textbf{training with (+) and without (-) in-domain data}, averaged over modals. Cross-validated averages and standard deviations of averages of per-modal scores.
	} 
	\label{tab:cross_domain}
\end{table}

\paragraph{Cross-Domain Results on \corpusName.}
We conduct a cross-domain experiment on \corpusName to determine the extent to which in-domain training data is necessary for classifying modal verbs in different scientific communities.
Since some modal-domain combinations have rather little data, in this experiment, we split \corpusName into six folds and use each fold once for testing.
We use four of the remaining five folds for training and one for early stopping.

\tref{cross_domain} reports the cross-validated averages and standard deviations of averages of per-modal \macrofscore and accuracy to show the overall effect of in-domain data.
Models trained on other (scientific) domains work well on unseen domains, as the performance does not decrease substantially when training without in-domain data.
As one would expect, domain-specific data usually leads to improvements, especially for domains in which a specific modal has a visibly different label distribution (see \fref{label_dist_genre_modal}), e.g., cross-validated \macrofscore for \textit{could} on CL increased by around 18 points.
For other modal-domain combinations, gains were less distinct or sometimes non-existent, and cross-validated scores had a high variance.
On average, standard deviation of accuracy was 2.5 and 2.7 for with and without in-domain data, respectively.
For \macrofscore, standard deviation was 10.7 when training with in-domain data and 11.3 when training without in-domain data.

In sum, we expect classifiers trained on \corpusName to also generalize to new scientific domains to some extent.
For optimal performance, adding in-domain data is beneficial in most cases.

\subsection{Transfer from GME to \corpusName}%
\label{sec:transfer-experiment}
In this experiment, we show that functions of modal verbs in scientific text cannot be determined simply using existing datasets.
We train a model only on an out-of-genre resource (GME in the version published by Pyatkin21).\footnote{We thank the anonymous reviewers for proposing this interesting experiment.} 
We train on \textbf{GME\textsubscript{T}}, i.e., all instances from GME (including the test set) that cover \corpusName's set of modal verbs using mapped labels as shown \tref{mapping_table}.
Resolving GME's \textit{State of Knowledge} into \labelInference and \labelUncertainty and \textit{State of the World} into \labelFeasibility and \labelOptions would require a manual re-annotation.
We map \labelDeontic to Pyatkin21's supertype \textit{Priority}.

\begin{table}[t]
    \centering
    \footnotesize
    \begin{tabular}{rl}
    \toprule
         \labelFeasibility, \labelOptions & \textit{State of the World} \\ \midrule
         \labelCapability, \labelRhetorical &  \textit{State of the Agent}\\\midrule
          \labelUncertainty, \labelInference & \textit{State of the Knowledge}\\\midrule
          \labelDeontic & \textit{Priority} (\textit{Desires+Wishes},\\
          & \textit{Plans+Goals}, \textit{Rules+Norms})\\
          \bottomrule
    \end{tabular}
    \caption{\textbf{Transfer experiment: Mapping} between GME (Pyatkin21) and \corpusName schemes.}
    \label{tab:mapping_table}
\end{table}

\begin{table}[t]
	\centering
	\footnotesize
	\setlength{\tabcolsep}{3pt}
	\renewcommand{\arraystretch}{1.1}
	\begin{tabular}{l|cccccc}
		\toprule
		& \textbf{can} & \textbf{could} & \textbf{may} & \textbf{might} & \textbf{must} & \textbf{should}\\
		\midrule
		Maj\tiny{GME\textsubscript{T}} & 33.5 & 19.7 & 15.9 & 30.7 & 30.2 & 28.9\\[-0.15cm]
		& \tiny{$\pm0.0$} & \tiny{$\pm0.1$} & \tiny{$\pm0.0$} & \tiny{$\pm0.0$} & \tiny{$\pm0.0$} & \tiny{$\pm0.0$}\\
		SB\tiny{CLS, modal;} & 56.1 & 43.7 & 19.8 & 30.2 & 33.9 & 39.8\\[-0.15cm]
		\hspace*{5mm}\tiny{GME\textsubscript{T}}& \tiny{$\pm7.7$} & \tiny{$\pm6.5$} & \tiny{$\pm4.4$} & \tiny{$\pm4.0$} & \tiny{$\pm7.7$} & \tiny{$\pm6.2$}\\
		SB\tiny{CLS, modal;} & 82.9 & 78.1 & \textbf{43.7} & 63.4 & 69.2 & 76.9\\[-0.15cm]
        \hspace*{5mm}\tiny{\corpusName-small}& \tiny{$\pm0.8$} & \tiny{$\pm1.7$} & \tiny{$\pm1.2$} & \tiny{$\pm5.0$} & \tiny{$\pm6.3$} & \tiny{$\pm11.6$}\\
		SB\tiny{CLS, modal;} & \textbf{84.3} & \textbf{79.6} & 43.1 & \textbf{69.9} & \textbf{74.6} & \textbf{84.1}\\[-0.15cm]
		\hspace*{5mm}\tiny{\corpusName}& \tiny{$\pm0.8$} & \tiny{$\pm1.8$} & \tiny{$\pm1.1$} & \tiny{$\pm1.1$} & \tiny{$\pm3.8$} & \tiny{$\pm2.4$}\\
		\bottomrule
	\end{tabular}%
	\caption{\textbf{Transfer experiment: Macro \fscore} on \textit{mapped} test set of \corpusName.
	} %
	\label{tab:evaluation_transfer_small}
\end{table}

GME\textsubscript{T} consists of 1276 instances, of which 370/238/139/61/196/272 are instances of \textit{can}, \textit{could}, \textit{may}, \textit{might}, \textit{must}, and \textit{should}, respectively. %
We train and evaluate all models in this experiment using the mapped annotation scheme, using the SB{\tiny CLS,modal} architecture with sigmoid heads.
For hyperparameter tuning and evaluation, we perform the steps as described in \aref{experiments-appendix} on GME\textsubscript{T} with five randomly induced folds (\textbf{SB{\tiny CLS,modal; GME\textsubscript{T}}}).
\textbf{SB{\tiny CLS,modal; \corpusName}} is SB{\tiny CLS,modal} trained on \corpusName with \textit{mapped labels}.
\textbf{SB{\tiny CLS,modal; \corpusName-small}} is trained on a randomly downsampled version of \corpusName to account for the notably larger size of \corpusName compared to GME\textsubscript{T}.
We approximately proportionally randomly downsample each \corpusName fold (with mapped labels) to get \textbf{\corpusName-small}, which has exactly the same number of instances as GME\textsubscript{T}.

\tref{evaluation_transfer} shows the results of our transfer experiment.
SB{\tiny CLS,modal; GME\textsubscript{T}} learns more than just the majority baseline \textbf{Maj{\tiny GME\textsubscript{T}}} (except for \textit{might}, for which GME contains little data), but clearly lags behind the models trained on \corpusName in both \macrofscore and accuracy, with average \macrofscore being between 23.9 and 37.1 points lower than models trained on \corpusName-small.
A related experiment (reported in \aref{cross-genre-multitasking}) using prior corpora of annotated modals in multi-task objectives confirmed the limited amount of transferability.
Hence, genre-specific data is clearly required for classifying functions of modal verbs in scientific discourse, demonstrating the value of \corpusName.

\begin{table}[t]
	\centering
	\footnotesize
	\setlength{\tabcolsep}{1pt}	
	\begin{tabular}{l|cccccc}
		\toprule
		\textbf{Macro \fscore}& \textbf{can} & \textbf{could} & \textbf{may} & \textbf{might} & \textbf{must} & \textbf{should}\\
		\midrule
		Maj\tiny{GME\textsubscript{T}} & 33.5 & 19.7 & 15.9 & 30.7 & 30.2 & 28.9\\[-0.15cm]
		& \tiny{$\pm0.0$} & \tiny{$\pm0.1$} & \tiny{$\pm0.0$} & \tiny{$\pm0.0$} & \tiny{$\pm0.0$} & \tiny{$\pm0.0$}\\
		SB\tiny{CLS, modal;GME\textsubscript{T}} & 56.1 & 43.7 & 19.8 & 30.2 & 33.9 & 39.8\\[-0.15cm]
		& \tiny{$\pm7.7$} & \tiny{$\pm6.5$} & \tiny{$\pm4.4$} & \tiny{$\pm4.0$} & \tiny{$\pm7.7$} & \tiny{$\pm6.2$}\\
		SB\tiny{CLS, modal;\corpusName-small} & 82.9 & 78.1 & \textbf{43.7} & 63.4 & 69.2 & 76.9\\[-0.15cm]
        & \tiny{$\pm0.8$} & \tiny{$\pm1.7$} & \tiny{$\pm1.2$} & \tiny{$\pm5.0$} & \tiny{$\pm6.3$} & \tiny{$\pm11.6$}\\
		SB\tiny{CLS, modal;\corpusName} & \textbf{84.3} & \textbf{79.6} & 43.1 & \textbf{69.9} & \textbf{74.6} & \textbf{84.1}\\[-0.15cm]
		& \tiny{$\pm0.8$} & \tiny{$\pm1.8$} & \tiny{$\pm1.1$} & \tiny{$\pm1.1$} & \tiny{$\pm3.8$} & \tiny{$\pm2.4$}\\
		
		\midrule
		\textbf{Accuracy}\\
		\midrule
		Maj\tiny{GME\textsubscript{T}} & 68.7 & 63.4 & 67.0 & 85.7 & 88.6 & 85.9\\[-0.15cm]
		& \tiny{$\pm0.0$} & \tiny{$\pm0.2$} & \tiny{$\pm0.0$} & \tiny{$\pm0.0$} & \tiny{$\pm0.0$} & \tiny{$\pm0.0$}\\
		SB\tiny{CLS, modal;GME\textsubscript{T}} & 76.1 & 70.1 & 68.9 & 74.7 & 88.9 & 85.6\\[-0.15cm]
		& \tiny{$\pm2.6$} & \tiny{$\pm2.7$} & \tiny{$\pm2.1$} & \tiny{$\pm10.5$} & \tiny{$\pm1.7$} & \tiny{$\pm2.9$}\\
		SB\tiny{CLS, modal;\corpusName-small} & 90.1 & 86.1 & \textbf{79.6} & 85.9 & 92.1 & 92.5\\[-0.15cm]
         & \tiny{$\pm0.3$} & \tiny{$\pm1.1$} & \tiny{$\pm1.0$} & \tiny{$\pm1.1$} & \tiny{$\pm0.7$} & \tiny{$\pm1.2$}\\
		SB\tiny{CLS, modal;\corpusName} & \textbf{91.0} & 87.0 & \textbf{79.6} & \textbf{88.4} & \textbf{93.4} & \textbf{94.3}\\[-0.15cm]
		& \tiny{$\pm0.5$} & \tiny{$\pm1.2$} & \tiny{$\pm0.9$} & \tiny{$\pm0.7$} & \tiny{$\pm0.7$} & \tiny{$\pm0.6$}\\
		\bottomrule
	\end{tabular}%
	\caption{\textbf{Transfer experiment: } on \textit{mapped} test set of \corpusName.
	} %
	\label{tab:evaluation_transfer}
\end{table}

\section{Conclusion and Outlook}

In this paper, we have introduced a new large-scale dataset of scientific text annotated for functions of modal verbs.
Our corpus and computational studies reveal differences and similarities in modal usage across genres and domains.
We have shown that neural classification is robust across scientific domains, but also that annotated scientific text is essential for good performance.
To sum up, our paper lays the groundwork for informed IE from sentences containing modals in scientific texts, e.g., distinguishing speculations from capabilities attributed to a method or device.

\textbf{Future work.}
Our experiments on \corpusName point to various next steps, e.g., identifying domain-adaptation methods that more effectively leverage annotations across domains, genres, or even languages, or developing data-augmentation techniques targeted to scientific text.
Another next step is to integrate our methods for generating metadata for facts into IE systems.
Existing open IE systems do not handle the meaning of modal verbs adequately.
As an outlook, in \aref{openie}, we outline how this could be improved using a classifier trained on \corpusName.

\clearpage
\section*{Acknowledgements}
We thank Alex Fraser, Alexis Palmer, Jannik Strötgen, Heike Adel, and the anonymous reviewers for their valuable comments.
Thanks also go to Daria Stepanova for a very helpful discussion on knowledge representation.
We thank our annotators Sherry Tan, Prisca Piccirilli, Johannes Hingerl, Federico Tomazic, and Anika Maruscyk for their dedication to the project and insightful discussions.
We also thank Josef Ruppenhofer for answering our questions on the Modalia annotation scheme, and Valentina Pyatkin and Shoval Sadde for answering our clarification questions on their models and experiments.

\section*{Limitations}
\textit{Closed class of targets.}
Our work is limited to a closed class of linguistic expressions (modal verbs).
Such approaches are sometimes seen as \enquote{too narrow} to be of interest to the NLP community.
However, we argue that examining components of language understanding in detail will ultimately point to relevant research directions.
In addition, as we have shown, modal verbs are a very common phenomenon, occurring in about every tenth sentence in scientific text.
Nevertheless, we admit that a limitation of our study is the focus on a closed set of verbs in the English language.
Future work might generalize our ideas to a more open class of targets (yet, it is a challenge to come up with a well-defined selection).

\textit{Limited data for minority classes.}
For some categories, data is limited due to the difficulty of data collection (we can only sample for modal verbs, not for labels).
We have already enriched the dataset by a second annotation round, further data collection is unfortunately infeasible in the context of our project.

\textit{Applications.}
Our study provides the first steps (an annotated dataset, a corpus-linguistic study and NLP models) of research into the computational modeling of modal verbs in scientific text.
Our distinctions intuitively should be of high relevance to processing and mining scientific text.
Besides the case study on why the distinctions matter for open information extraction (IE) and practical suggestions for incorporating them into existing Open IE systems in \aref{openie}, demonstrating the usefulness of our work on existing scientific relation extraction datasets (which unfortunately do not commonly mark the \enquote{information status} of the annotated relations) is beyond the scope of this paper (but planned future work).

\section*{Ethics Statement}
The corpus described in this work consists of open-access scientific articles annotated with several categories.
The annotators involved in the project gave their explicit consent to the publication of their annotations and were compensated substantially above the minimum wage in our country.

\bibliography{anthology,custom}
\bibliographystyle{acl_natbib}

\clearpage
\section*{Supplementary Material}
\appendix

\section{Comparison to Existing Annotation Schemes for Modal Senses}
\label{sec:scheme-comparison}
Table \ref{tab:scheme_comparison} classifies a set of utterances according to our, RR12's and Rubin13's schemes (according to our interpretation of their guidelines).\footnote{To facilitate comparison with Pyatkin21, we also added their mapping to the Rubin13 scheme to the table.}
These two works inspired ours, but with the aim of knowledge graph construction in mind, we tailored an annotation scheme making explicit the various pragmatic and rhetorical reasons for using modals in scientific writing.
We thereby follow \citet{moon-etal-2016-selective}, who argue that \enquote{not everything about modal auxiliary meaning can be represented at once,} and that \enquote{it is important to focus on the parts of modal auxiliary meaning that most directly impact an automated learner.}
While we fully agree with the linguistic classification of the examples by RR12 and Rubin13, we found certain sub-distinctions to be essential for understanding modal usage in the scientific context, and designed our annotation scheme for \textit{functions} of modals accordingly, intentionally conflating what is traditionally treated separately as \textit{modal sense disambiguation} and \textit{veridicity} \citep{karttunen2005veridicity} from the author's point of view.

The definition of Rubin13's label \textit{Circumstantial}, focusing less on dispositions rather than on abilities in particular circumstances \citep{fintel2006modality}, is closer to our \labelFeasibility, which could be interpreted as an ability of the actor given the circumstances (but sometimes overlaps with internal properties of the object under discussion).
Conversely, we do not distinguish personal wishes and goals as in Rubin13.
The label \textit{deontic} for \textit{can} of RR12 falls under our label \labelOptions if options are introduced, and maps to our \labelDeontic otherwise.
Within the \textit{epistemic} notion, we further distinguish whether a statement is derived from other facts (\labelInference) or whether an author \textbf{\textit{speculates}} (both labels may apply at the same time).
As some usages of modal verbs in scientific writing are rather conventional, we introduce the label \labelRhetorical.

\section{Further Corpus Statistics}
\label{sec:further-corpus-stats}

\subsection{Impact of Negation}
Analyzing all negated modal verb constructions, we found only two instances where negation affects the modality label.
For example, \enquote{Submarine volcanism alone \dul{cannot} be the sole driving mechanism for OAEs} is labeled with \labelCapability when ignoring the negation.
Otherwise, this becomes an \labelInference.

\subsection{Comparison of Label Distributions of \corpusName, MASC, and Modalia\textsubscript{M}}
\label{sec:masc-modalia-stats}

\begin{figure*}[t]
	\centering
	\begin{subfigure}{.38\textwidth}
		\includegraphics[height=25mm]{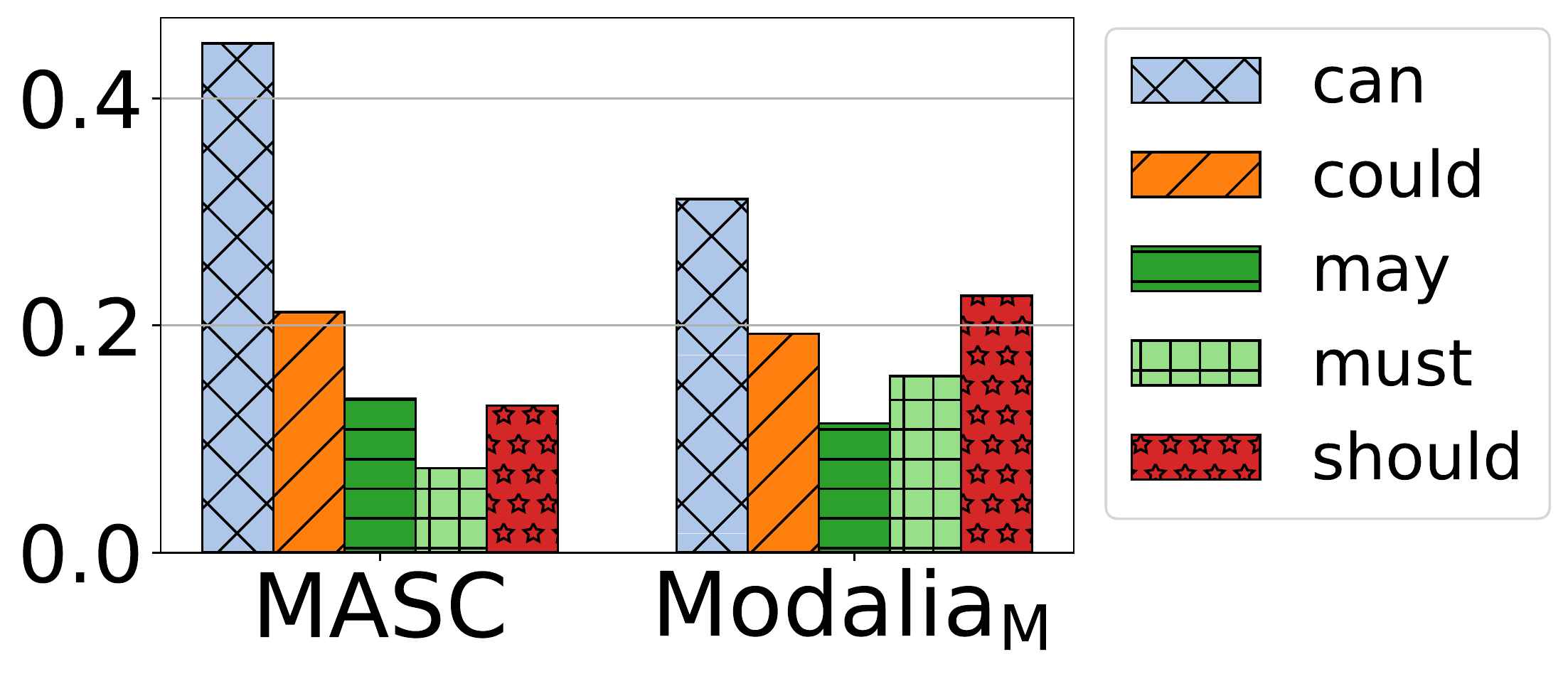}  
		\caption{Distributions of modals}
		\label{fig:masc_modal_dists}
	\end{subfigure}
	\begin{subfigure}{.02\textwidth}
		\footnotesize
		\rotatebox{90}{\hspace*{10mm}Percentage}
	\end{subfigure}
	\begin{subfigure}{.08\textwidth}
		\centering
		\includegraphics[height=28mm]{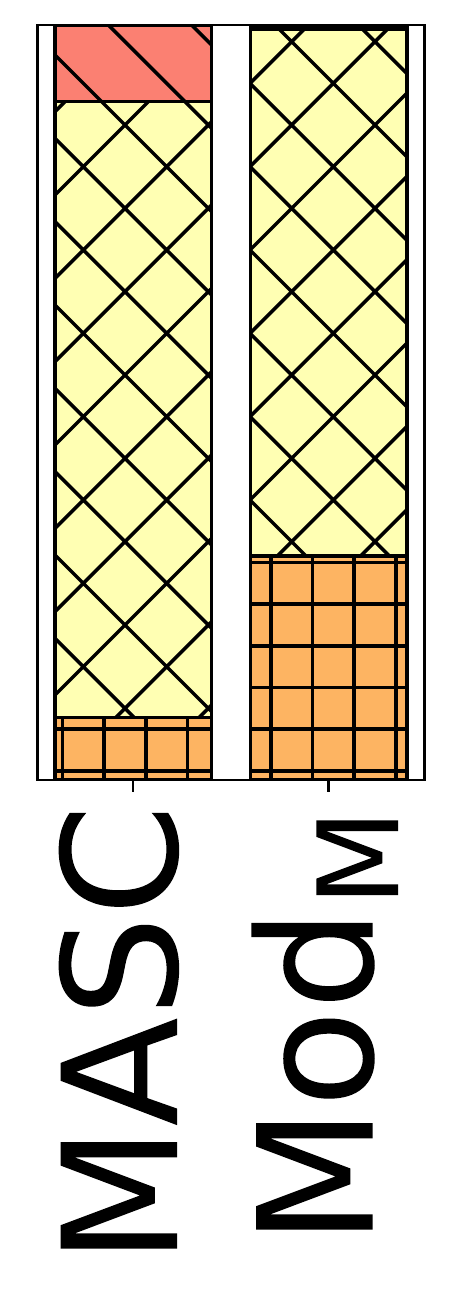}
		\caption{can}
	\end{subfigure}
	\begin{subfigure}{.08\textwidth}
		\centering
		\includegraphics[height=28mm]{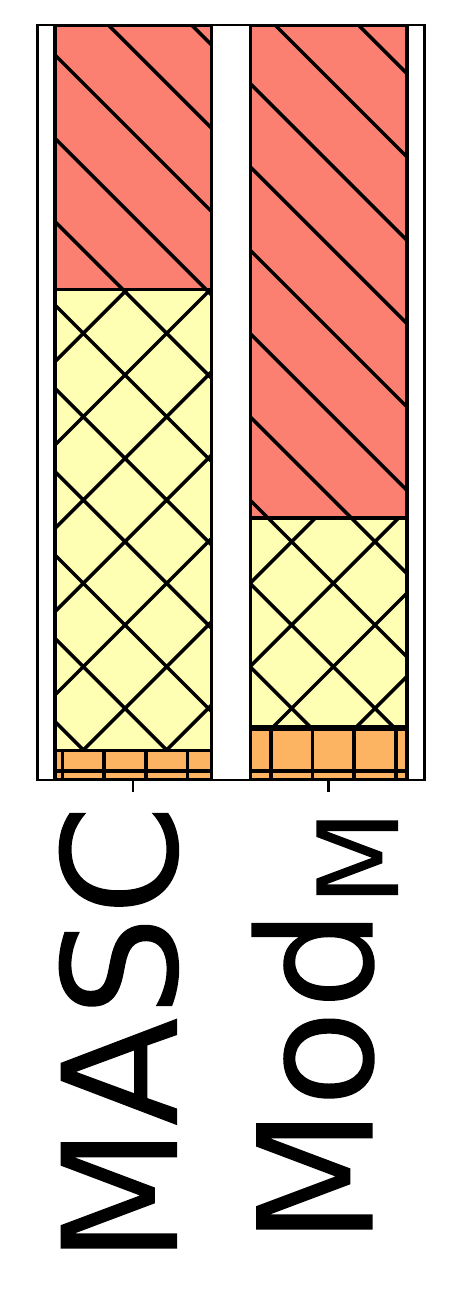}  
		\caption{could}
	\end{subfigure}
	\begin{subfigure}{.08\textwidth}
		\centering
		\includegraphics[height=28mm]{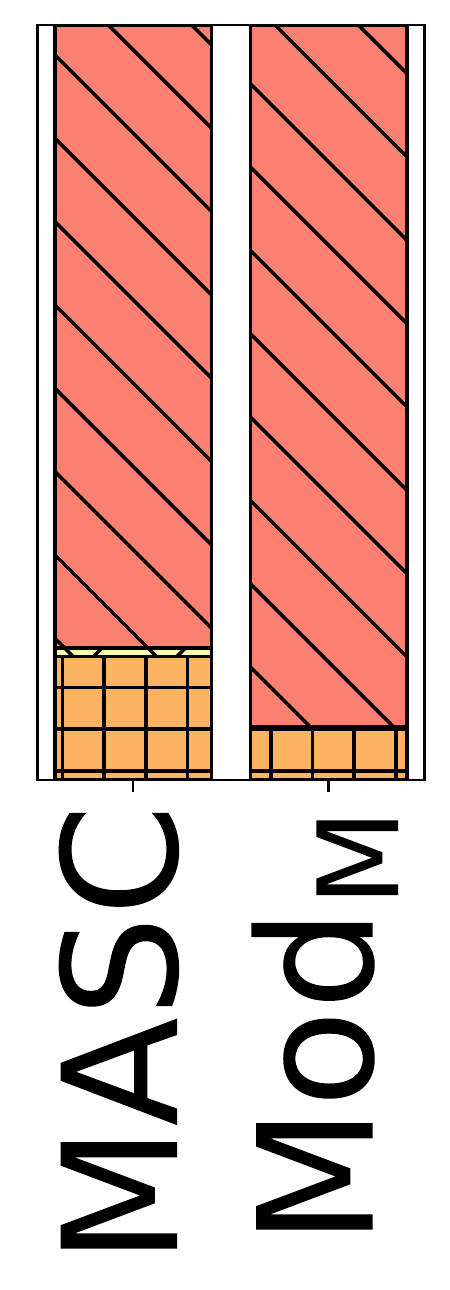}  
		\caption{may}
	\end{subfigure}
	\begin{subfigure}{.08\textwidth}
		\centering
		\includegraphics[height=28mm]{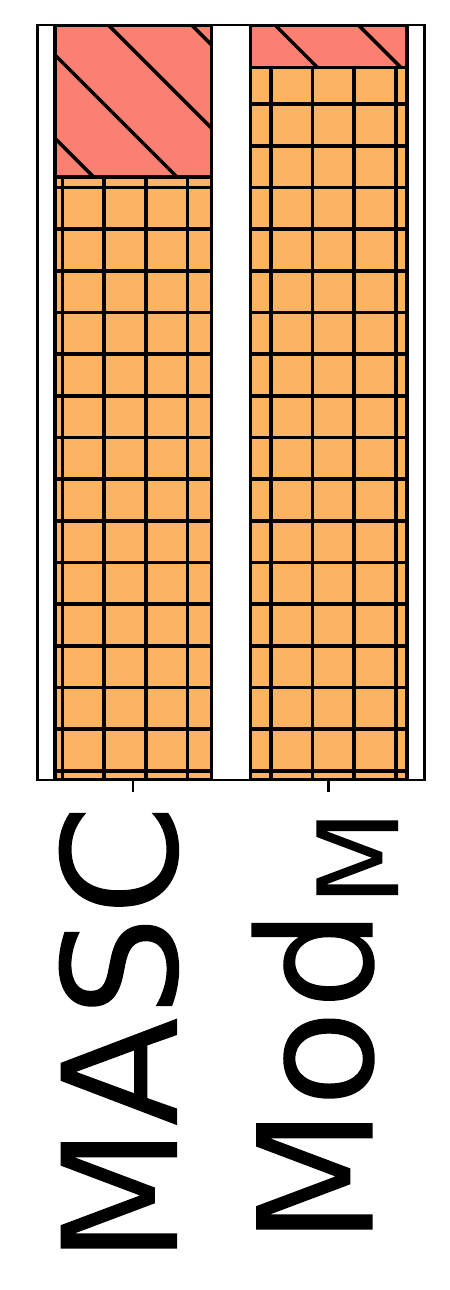}  
		\caption{must}
	\end{subfigure}
	\begin{subfigure}{.24\textwidth}
		\includegraphics[height=28mm]{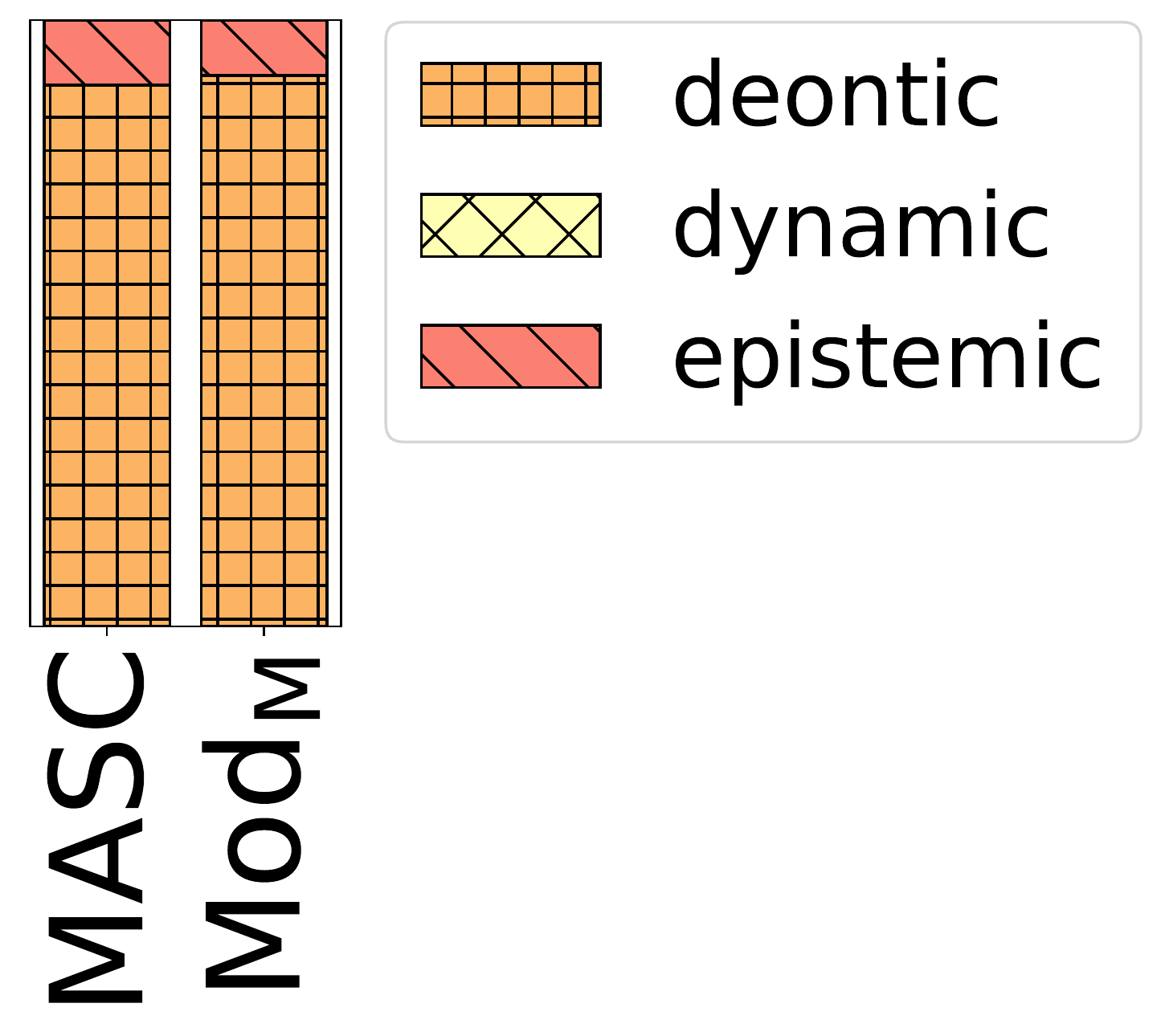}  
		\caption{should\hspace*{2cm}}
	\end{subfigure}
	\caption{\textbf{MASC and Modalia\textsubscript{M}}: Modal distributions and label distributions by modal verb.} %
	\label{fig:label_dists_masc_modalia}
\end{figure*}

The distribution of modal functions and sense differs between corpora and genres (academic writing vs. news).
Comparing \fref{label_dist_genre_modal} and \fref{label_dists_masc_modalia}, we note several differences.
The most frequent modal in all genres is \textit{can}, but it is much more frequent in CL, CS, and MS.
For \textit{can} and \textit{could}, \textit{dynamic}/\labelFeasibility/\labelCapability uses are predominant, with the exception of Modalia\textsubscript{M}, where the majority class of \textit{could} is \textit{epistemic}.
\textit{Can} and \textit{could} are not used in the \textit{deontic} sense in \corpusName; their \textit{epistemic} uses are all related to \labelUncertainty.

\section{Annotation Guidelines}
\label{sec:details-annotation-scheme}
In this section, we describe our annotation guidelines for marking up modal verbs in scientific publications with regard to whether they are used for particular rhetorical, semantic or pragmatic reasons \textit{as they were presented to the annotators}.
Depending on the context, modal verbs can modify a sentence's propositional content such that uncertainty about the truth of the proposition is implied (e.g., \enquote{X \textit{is} the cause for Y} vs. \enquote{X \textit{may} be the cause for Y}), but in other circumstances, they simply indicate properties or capabilities (e.g., \enquote{X \textit{can} float}).
Our goal is to provide information about the functions of modal verbs in our corpus that then can be used in a preprocessing step for information extraction.
For example, when disregarding a modal's contribution to the discourse, when processing \enquote{X \textit{can} float}, the relation \texttt{float(X)} may be extracted, but it should be flagged somehow as the sentence does not state that X is currently floating or that it always floats.
In contrast, adding has\_capability(X, float) to our knowledge base is desirable.

We consider \textit{can, could, must, should, may, might} as well as their negated forms for annotation.
They are pre-marked in the corpus to ensure that no modal verb is overlooked.
Our annotation scheme is based on the observation that it is not always possible to assign exactly one type to every instance.
We decided to follow a feature-based annotation approach in which a modal verb is represented by features that do or do not apply.
Our feature sets reflects the range of functions a modal verb can have, i.e., the meaning that it adds to the sentence (e.g., capability), or the rhetorical or pragmatic reason for using it (e.g., uncertainty).
To determine which features apply, annotators are asked to think of the sentence without the modal verb first, and then observe how the meaning has changed when adding the modal.
Selecting the features accordingly means determining the reason(s) for which the author uses the respective modal verb.
The next section explains the set of set of functions in our scheme.

It is important to note that we annotate \textit{the author's} intentions and understanding, not the reader's (which might be based on additional context or knowledge).
However, for making the judgment of why an author uses a modal verb in a particular context, annotators are of course asked to consider the broader context.

\subsection{Functions of Modal Verbs}

\begin{description}
\setlength{\itemsep}{1pt}
\setlength{\parskip}{0pt}
\setlength{\parsep}{0pt}
	\item[\labelFeasibility:] We use this feature when it is possible for an \dul{external actor} to do or achieve something and indicating this is the reason why the modal verb is used.
\labelFeasibility can be seen as general possibility involving some external actor, e.g., a human agent.
\begin{example}
\textit{Several supercapacitors \textbf{can} be integrated and connected in series.}\\
It is possible for somebody to integrate and connect supercapacitors, it needs a
human agent to do it. The focus is not on an internal capability of the supercapacitors here.
\end{example}	

	\item[\labelCapability:]
\labelCapability is annotated if the modal verb is used to express that something has a certain \dul{property, ability or capacity}.
\citet{ruppenhofer-rehbein-2012-yes} have a corresponding category named \textit{dynamic}.
We mark up \labelCapability only in cases where the modal verb is used to convey information about an \underline{intrinsic} property of an entity.
\begin{example}
\textit{The device \textbf{can} light up a redlight-emitting diode and works well.}\\
The device has the ability to light, the device is able to light up, being able to light up is an inherent property of the device.
\end{example}
\begin{example}
\textit{ They hope the government \textbf{can} introduce a new law}.\\
The government is able to introduce a new law. Therefore \labelCapability is marked up. Note that even if the sense of the utterance is \enquote{it is desirable that the government introduces a new law,} \labelDeontic doesn't apply here as the desire is expressed by \enquote{hope} and not by the clause containing the modal verb.
\end{example}
	
	\item [\labelInference:]
	This feature covers cases in which an author states that she \dul{inferred} something based on some given information. 
\labelInference corresponds to the category \textit{epistemic} in previous work, as it also applies if the author draws a conclusion based on some information.
\labelInference applies especially when the author predicts something, e.g., based on computational results, experimental outcomes, or empirical knowledge.
In order to correctly identify this feature, usually a broader context needs to be taken into consideration and domain knowledge is sometimes crucial.
\begin{example}
\textit{The maximum power density was measured to 0.350 mW cm$^2$. Therefore, it \textbf{must} be the case that the open-circuit voltage reaches at least 1 V.}\\
Based on the measurement of the power density, the author infers that the open-circuit voltage is 1 V.
\end{example}
\begin{example}
\textit{(According to these calculations...) The three lowest-energy isomers of C60O3 \textbf{should} exist in equilibrium at room temperature by using a modified and extended Hűckel method.}\\
The author predicts that these isomers exist in equilibrium at room temperature, based on some calculations.
\end{example}
	
	\item [\labelUncertainty:]
	\labelUncertainty is used when the \dul{truth} value of an utterance is not clear according to the author.
Note that we annotate this feature only in cases where \labelFeasibility or \labelCapability are not clearly the predominant readings, and use both features only if a speculation reading is really predominant.

\begin{example}
	\textit{This problem \textbf{might} be mitigated by using better semantic-based retrieval model.}\\
	Here, we label both \labelFeasibility and \labelUncertainty.
	Consider replacing \textit{might} with \textit{can}: then, the \labelFeasibility is clear, but no \labelUncertainty is involved, which is the author's reason for choosing \textit{might} instead.
\end{example}

	\item [\labelOptions:] 
	\labelOptions is marked up when the author uses the modal verb to enumerate some \underline{\smash{potential options}}.
\begin{example}
\textit{The real shielding \textbf{can} of course be different.}\\
A different shielding may be used; potentially the shielding is different, but it can also stay the same.
Note that \enquote{being different} is not a property, hence \labelCapability or \labelFeasibility wouldn't fit here.
\end{example}
\begin{example}
\textit{Grounding this in our example, w1 \textbf{may} represent breakfast, w2 pancakes, and w4 hashbrowns.}\\
Breakfast, pancakes and hashbrowns are options for w1, w2 and w4.
\end{example}
\begin{example}
	\textit{
This process \textbf{can} last from several hours to a few days depending on the applied temperature.}\\
The reason for using the modal verb here is mostly to convey the uncertainty about the duration, it does not describe a capability of the process.
\end{example}
\begin{example}
	\textit{We showed that combining a model based on minimal units with phrase-based decoding \textbf{can} improve both search accuracy and translation quality.}\\
	In this case, we label both \labelCapability and \labelOptions, as the sentence both indicates a capability of the combination method, but at the same time could be read as a hedging term (i.e., improvements occur only in certain circumstances).
\end{example}

	\item [\labelDeontic:] 
	\labelDeontic is selected if the author uses the modal verb to express a \dul{desire}, i.e., how the world should be like, to express a \dul{requirement} for something, e.g., an experiment, or to state an \dul{obligation}.
\begin{example}
\textit{Our daily life requires matchable energy storage devices, which \textbf{should} have the capability to endure high-level strains.}\\
It is desirable that energy storage devices have the capability to endure high-level
	strains.
\end{example}
\begin{example}
	\textit{A GCR proton at the maximum latitude of the ISS \textbf{must} have at least about 150 MeV to reach the Station.}\\
	It is required that a GCR proton has at least about 150 MeV to reach the station.
\end{example}
\begin{example} \textit{Temperature \textbf{should} be treated as a concept.}\\
The author prescribes that temperature is treated as a concept, it is necessary that temperature is treated as a concept.
\end{example}
	
	\item [\labelRhetorical:] In some contexts, modal verbs are used because of \underline{conventions} and there is no substantial semantic need for doing so.\footnote{In several cases that we observed, we also felt that they corresponded to over-use of modal verbs by non-native English speakers (though they are not syntactically wrong, just unnecessary to some extent).}
We annotate this cases with \labelRhetorical. 

\begin{example}\textit{It \textbf{can} be seen in Figure 1 that...}\\
Can simply be stated at `In Figure 1 you see ...''
If annotators feel that \labelFeasibility or \labelCapability are also strongly present in such a case, they may select these features in addition.
\end{example}

\begin{example}
\textit{Value: <first part of the definition> The value \textbf{can} also be described via <second part of the definition>.}\\
A value is defined as (first part of the definition) and also as (second part of the definition). (In this example, \labelUncertainty and \labelFeasibility are also applicable.)
\end{example}

\item[Other:] This label is used if none of the above features apply. Please extract those sentences and explain why you couldn't decide for a predefined feature. Also think about whether you have a tendency towards one or more features but there is something that has to be captured  in our scheme in additional. \textit{This was used during annotation scheme development.}
\end{description}

\subsection{Additional examples: \labelFeasibility vs. \labelCapability}

As stated above, features are not mutually exclusive. Sometimes, multiple readings/interpretations may be possible.
Under certain circumstances annotators are asked to select multiple features.
In this section, we show some not-so-clear-cut examples to complement the above guidelines, which work with mostly clear examples.

If we want to annotate \labelFeasibility or \labelCapability but it is hard to decide which of both features apply, we follow the following guidelines.

We annotate both features when there is an external actor (e.g., a human) involved, but if it can also be interpreted as describing a particular internal property of the referent of the subject.
The referent has this property already before an external actor is involved.

\begin{example}
\textit{This simpler distribution Q \textbf{can} be viewed as an approximation to P.}
\\
\labelFeasibility: A human agent views Q as an approximation to P.\\
\labelCapability: Without a human agent viewing Q, Q is still an approximation to P, P has the property of being an approximation to P in general without somebody actually viewing it.
\end{example}

\begin{example}
\textit{For instance, despite graphene, the band gaps of silicone \textbf{can} be opened and tuned when exposed to an external electric field.}\\
\labelFeasibility:A human agent opens the band gaps of silicone.\\
\labelCapability: some materials have the property of having openable band gaps, it is always possible to open band gaps of silicone under this circumstances.\\
\end{example}

Whenever there is a \dul{human actor} involved, we mark up \labelFeasibility even if the sentence includes a passive construction which could indicate a \labelCapability.
\labelCapability \dul{and} \labelFeasibility are only used at the same time if the modal verb is used to signal an intrinsic property (band gaps \textit{of} silicone can be opened) \dul{and} an external actor is involved.
We do not mark up \labelCapability if \labelFeasibility applies but there isn't a general property. In this case an external actor has to do something first. As a consequence, some entity has a capability.

\begin{example}
\textit{The resulting expression combines similarity terms which \textbf{can} be divided into two groups.}
\\ \labelFeasibility: An human actor is needed to divide the terms into groups.
Being dividable is not an intrinsic, common property of these terms. \labelFeasibility is the strongest modal function in this utterance. The modal verb is not used to convey an information about a \labelCapability, as it is not an intrinsic property of similarity terms that they can be divided (we consider this to be an artifact of their being grouped).
\end{example}

\begin{example}
\textit{Similar symmetry \textbf{can} be achieved with the following factorization.}\\
\labelFeasibility: it needs a human agent to achieve something.\\
The modal verb is not used to indicate an intrinsic property of being achievable. The sense of the utterance is that somebody, i.e., the author, achieves similar symmetry with a certain formula that they mention. Even if no human actor is explicitly mentioned in the text due to a passive construction, only \labelFeasibility may apply.
\end{example}

\begin{example}
	\textit{Word vectors \textbf{can} be trained directly on a new corpus.}\\
	\labelFeasibility: It is possible for somebody to train some word vectors on a new corpus.\\
	Word vectors cannot be trained directly on a new corpus in general, not all word vectors are trainable on a new corpus, so we don't annotate \labelCapability.\\
\end{example}

When it is clearly possible for an entity to have a property but this doesn't apply in general, we still use \labelCapability, but possibly \labelUncertainty in addition.

\begin{example}
	\textit{graphene aerogels with ... \textbf{can} present superelasticity.}\\
	\labelCapability: some of these aerogels have this property, it is possible for aerogels to
	have this property.
	\labelUncertainty: It is uncertain whether each aerogel has this property, only some of them may present superelasticity, or aerogels have this property only under particular circumstances.
\end{example}
\section{Experimental Studies}
\label{sec:experiments-appendix}

\subsection{Hyperparameters}
\label{sec:hyperparameters}
\begin{table}[]
    \centering
    \footnotesize
    \begin{tabular}{c|c|c}
         Hyperparameter & CNN & SB \\
         \toprule
         Learning rate & $1e-3,5e-3,$ &  $5e-4, 3e-5, $\\
         & $1e-4, 5e-4,$ & $5e-5$\\
         & $3e-5,5e-5$ & \\
         \midrule
         \# warm-up epochs & N/A & $1,2$ \\
         \midrule
         Batch size & $8,16,32$ & $8,16,32$ \\
         \midrule
         Dropout & $0.1, 0.5$ & N/A \\
    \end{tabular}
    \caption{\textbf{Hyperparameter values} searched during hyperparameter selection for CNN and SB.}
    \label{tab:grid_search}
\end{table}
This section describes the hyperparameter tuning for our main experiments.
For CNN and SB, we tune learning rates, batch sizes, dropout probabilities (only CNN) and learning rate warm-up lengths (only SB) using grid search on the values shown in \tref{grid_search} as follows:
Similar to cross validation (CV), for each hyperparameter configuration, we train five models on 4 folds each for 10 epochs and use the respective remaining fold (\textit{validation fold}) for model selection.
For each of the five models, we average weighted \fscore scores (see \sref{eval_metrics}) on the validation fold across modal verbs.
We then choose the hyperparameter setting that performs best on average  across the different models.
The tuned batch sizes and learning rates are $32$ and $5^{-3}$ (CNN), and $8$ and $3^{-5}$ (SB).
SB is warmed up for 2 epochs.
We use a dropout probability of $0.1$ in the output heads, and the Adam optimizer \citep{kingma2014adam} with a weight decay of $1^{-3}$ (CNN) respectively $0$ (SB).

\subsection{Training Details, Model Size, etc.}
All experiments were performed on a single Nvidia Tesla V100 GPU.
Training and testing the SB{\tiny CLS,modal} models in the 5-fold CV training setting used in the model architecture comparison experiment (cf. \tref{evaluation_modals_scientific_f1}) took 1.2 hours (for the entire experiment).

SciBERT has the same number of parameters as BERT-base, i.e., 110M. 
The linear layer we add on top of SciBERT in the SB{\tiny CLS,modal} has less than 11k parameters.

\subsection{Further Experiment Results}
\label{sec:further-results}
This section provides further experimental results, elaborating on the study described in \sref{mainresults}.

\tref{evaluation_modals_scientific_acc} provides accuracy scores for the models whose \fscore scores are reported in \tref{evaluation_modals_scientific_f1}.

\begin{table}[t]
	\centering
	\footnotesize
	\setlength{\tabcolsep}{2.5pt}	
	\begin{tabular}{l|cccccc}
		\toprule
		& \textbf{can} & \textbf{could} & \textbf{may} & \textbf{might} & \textbf{must} & \textbf{should}\\
		\midrule
		Maj & 85.0 & 78.1 & 80.7 & 91.5 & 93.5 & 92.0 \\
		CNN & 91.3 & 85.8 & 84.5 & 91.2 & 93.6 & 93.3\\
		SB\tiny{CLS} & 93.7 & 90.1 & 86.7 & 92.1 & \textbf{96.7} & 96.2\\
		SB\tiny{modal} & 94.2 & 90.2 & 86.6 & \textbf{92.6} & 96.6 & 96.7\\
		SB\tiny{CLS-mark} & \textbf{94.4} & 89.3 & 86.6 & 91.9 & 96.2 & 96.8\\
		SB\tiny{CLS, modal} & 94.3 & \textbf{90.8} & \textbf{87.0} & \textbf{92.6} & \textbf{96.7} & \textbf{96.9}\\
		BERT\tiny{CLS,modal} & 93.7 & 90.7 & 86.5 & 92.1 & 96.0 & \textbf{96.9}\\
		BERT-large\tiny{CLS,modal} & 94.3 & 89.6 & 86.0 & 92.3 & 96.5 & 96.8\\
		\bottomrule
	\end{tabular}%
	\caption{\textbf{Accuracy on test set of \corpusName}. Standard deviations are rather small, between 0 and 1.4.
	} 
	\label{tab:evaluation_modals_scientific_acc}
\end{table}

\subsection{Cross-Genre Multi-Tasking Experiment}
\label{sec:cross-genre-multitasking}
\begin{table}[t]
	\centering
	\footnotesize
	\setlength{\tabcolsep}{3pt}
	\begin{tabular}{l|cccccc}
		\toprule
		Train & \textbf{can} & \textbf{could} & \textbf{may} & \textbf{might} & \textbf{must} & \textbf{should}\\
		\midrule
		\corpusName & 77.4 & 73.7& 47.2 & \textbf{64.5} & \textbf{78.4} & 85.7\\[-0.15cm]
		& \tiny{$\pm1.0$} & \tiny{$\pm3.8$} & \tiny{$\pm1.1$} & \tiny{$\pm2.7$} & \tiny{$\pm1.1$} & \tiny{$\pm0.5$}\\
		+ EPOS & \textbf{78.4} & 69.6 & 49.4 & 64.0 & 74.9 & 86.1\\[-0.15cm]
		& \tiny{$\pm1.5$} & \tiny{$\pm3.9$} & \tiny{$\pm2.2$} & \tiny{$\pm2.4$} & \tiny{$\pm2.4$} & \tiny{$\pm2.5$}\\
		+ MASC & \textbf{78.4} & 72.2 & \textbf{51.5} & 63.2 & 75.8 & 84.4\\[-0.15cm]
		& \tiny{$\pm1.6$} & \tiny{$\pm1.3$} & \tiny{$\pm1.5$} & \tiny{$\pm3.2$} & \tiny{$\pm1.9$} & \tiny{$\pm1.0$}\\
		+ Modalia{\tiny M} & 76.6 & 71.5 & 49.4 & 59.9 & 75.6 & \textbf{86.4}\\[-0.15cm]
		& \tiny{$\pm1.5$} & \tiny{$\pm1.9$} & \tiny{$\pm3.3$} & \tiny{$\pm3.4$} & \tiny{$\pm4.3$} & \tiny{$\pm1.4$}\\
		+ GME & 76.7 & 70.5 & 47.0 & 62.3 & 70.8 & 84.1\\[-0.15cm]
		& \tiny{$\pm2.1$} & \tiny{$\pm4.2$} & \tiny{$\pm2.4$} & \tiny{$\pm6.0$} & \tiny{$\pm6.1$} & \tiny{$\pm1.6$}\\
		+ E/M/Mo & 77.0 & \textbf{73.9} & 50.0 & 62.8 & 77.2 & 85.6\\[-0.15cm]
		& \tiny{$\pm1.6$} & \tiny{$\pm1.7$} & \tiny{$\pm2.4$} & \tiny{$\pm3.3$} & \tiny{$\pm1.9$} & \tiny{$\pm0.6$}\\
		\bottomrule
	\end{tabular}%
	\caption{\textbf{Multi-task setup: Macro \fscore} on test set of \corpusName when co-training with other corpora. E/M/Mo: EPOS, MASC and Modalia{\tiny M} together.
	} 
	\label{tab:evaluation_modals_scientific_multi}
\end{table}

We investigate whether we can improve classification on \corpusName by using existing modal sense classification datasets as auxiliary tasks in training.
\tref{evaluation_modals_scientific_multi} shows the results of co-training with Modalia\textsubscript{M}, MASC, EPOS, and GME (see \sref{relwork}), and the first three at once.
On GME, we follow Pyatkin21's experiments and collapse \textit{Desires+Wishes} and \textit{Plans+Goals} to a \textit{Intentional}. %

The only verb where co-training leads to clear improvements is \textit{may}.
Here, it increases per-label \fscore scores (not reported in tables) for \labelUncertaintyShort, \labelOptionsShort, \labelFeasibilityShort (for the latter two except for GME), and \labelCapabilityShort (except for Modalia\textsubscript{M}).
For the other verbs, classification performance is similar (e.g., \textit{should}) or decreased (e.g., \textit{might}, which is only covered by GME, but just with few instances).
Thus, in line with the findings from the pure transfer experiment, using modal sense information from out-of-genre datasets for classifying modal verbs in scientific text is non-trivial.

\section{Case Study: Treatment of Modality in Open Information Extraction}
\label{sec:openie}

We now discuss how handling modal verbs in Open Information Extraction (OIE) systems may be improved using our classification scheme by adding interpretations instead of just pin-pointing modal verbs.
The same principles can be applied to relation extraction settings with predefined schemas when relations are rooted in a verbal argument structure.

\subsection{Analysis of Existing OIE Systems}
We run four popular recent OIE systems on sentences from \corpusName and perform a qualitative analysis of the results.
We find that the examined systems either have no specific mechanism for handling modality, or include modality information only in rather rudimentary ways.
\textbf{OpenIE4}\footnote{\href{https://knowitall.github.io/openie/}{knowitall.github.io/openie/}} \citep{christensen2011analysis,pal-mausam-2016-demonyms} and \textbf{OpenIE6}\footnote{\href{https://github.com/dair-iitd/openie6}{github.com/dair-iitd/openie6}} \citep{kolluru-etal-2020-openie6} extract information from sentences in the form of standard subject--relation--object triples, simply considering modals part of the predicate, e.g., a sentence such as \enquote{X \dul{may} influence Y} yields the extraction (X; may influence; Y).
\mbox{\textbf{RnnOIE}}\footnote{\href{https://demo.allennlp.org/open-information-extraction}{demo.allennlp.org/open-information-extraction}} \citep{stanovsky-etal-2018-supervised} generates a representation resembling Semantic Role Labeling (SRL), in which spans within the sentence are annotated to indicate the relation-evoking verb and its respective arguments, e.g., \texttt{[ARG0: X] [ARGM-MOD: may] [V: influence] [ARG1: Y].}
Within this representation, modal verbs are treated as a simple modifier of the relation verb (ARGM-MOD).
In sum, modals are extracted by all of these OIE systems, but their classification and interpretation is left to the downstream system.

\textbf{MinIE}\footnote{\href{https://github.com/uma-pi1/minie}{github.com/uma-pi1/minie}} \citep{gashteovski-etal-2017-minie} includes a notion of modality by adding a binary modality value (\textit{certainty}/\textit{possibility}) to each extracted triple.
In practice, we observe that the occurrence of virtually any modal in the input sentence results in the triple being assigned the \textit{possibility} label.
This means that sentences such as \enquote{X \dul{can} influence Y,}  \enquote{X \dul{should} influence Y,}  \enquote{X \dul{must} influence Y,} or  \enquote{X \dul{may} influence Y} are in effect all being treated as paraphrases. %
In sum, existing state-of-the-art OIE systems do not handle the meaning of modal verbs in a way that could inform downstream use.

\subsection{Discussion: Modality-informed Open IE}
In light of the weaknesses of existing systems, we now sketch an approach by which OIE systems could be extended to incorporate modality information, which could be generated by a classifier (as described in \sref{modeling}). %
As motivated by \fref{teaser-img}, we posit that there are two main ways in which modality information should be incorporated into extractions.
(For an overview, see also \tref{modality_ie_mapping}.)
First, we propose specific relation templates for the \labelCapability and \labelDeontic modalities: \textit{hasCapabilityTo\_*} for the former and \textit{isRequiredTo\_*} and \textit{isAllowedTo\_*} for the latter.
In a given extracted triple, these relation templates would be instantiated with the main verb of the extraction, e.g., \enquote{X \dul{can} influence Y} (\labelCapability) would yield \textit{(X, hasCapabilityTo\_influence, Y)}.\footnote{In an OWL-like ontology, these concretely instantiated predicates may then be considered subproperties of generic \textit{hasCapabilityTo} / \textit{isRequiredTo} / \textit{isAllowedTo} properties.}

Second, to cover cases modifying not only the relation but the entire fact, we propose the meta-property \textit{hasFactualityRating} (see also \fref{teaser-img}). %
This property could take the values \textit{speculation} (for \labelUncertainty), \textit{possible} (for \labelOptions and \labelFeasibility), \textit{inferred} (for \labelInference), and \textit{true} (for \labelRhetorical and as the default value of the property).
For example, the sentence \enquote{X \dul{might} influence Y} (\labelUncertainty) would yield \textit{(X, influence, Y)} with \textit{hasFactualityRating(speculation)}, whereas \enquote{These sandstones \dul{may} contain reworked material.} (\labelOptions), would lead to \textit{(sandstones, contain, reworked\_material)} with \textit{hasFactualityRating(possible)}.
Similar approaches to handling veridicality of utterances have for instance been proposed by \citet{de-marneffe-etal-2012-happen}.

We argue that such an approach would constitute an improvement over existing ways of handling modality in OIE.
Enabling the identification across surface representations is one aim of OIE systems.
Looking further ahead, explicitly disambiguating modal verbs as well as other constructions expressing the same meaning will result in a uniform representation.
For example, \textit{X can (\textbf{capability}) influence Y)} and \textit{X is able to influence Y} would both be retrieved by searching for \textit{hasCapability}, and \textit{X must (\textbf{deontic}) Y} and \textit{X has to Y} would be retrieved when searching for \textit{isRequiredTo}.
In addition, \textit{hasFactualityRating} properties of extracted triples will immediately clarify their factuality status, avoiding, e.g., erroneously taking speculation as fact.
Taken together, we have outlined a way to take OIE systems to the next level with regard to the treatment of modal verbs.

\begin{table}[t]
	\centering
	\footnotesize
	\begin{tabular}{ll}
		\toprule
		\textbf{Modal function} & \textbf{IE extraction(s)}\\
		\midrule
		\textbf{\textit{capability}} & \textit{hasCapabilityTo\_*}\\
		\textbf{\textit{deontic}} & \textit{isRequiredTo\_*} (must, should) / \\
	                              & \textit{isAllowedTo\_*} (other modals)\\
	                             \midrule
	   \textbf{\textit{feasibility}} & \textit{hasFactualityRating(possible)}\\
		\textbf{\textit{inference}} & \textit{hasFactualityRating(inferred)}\\
		\textbf{\textit{speculation}} & \textit{hasFactualityRating(speculation)}\\
		\textbf{\textit{options}} & \textit{hasFactualityRating(possible)}\\
		\textbf{\textit{rhetorical}} & \textit{hasFactualityRating(true)}\\
		\bottomrule
	\end{tabular}
	\caption{\textbf{Mapping modal functions to Open IE extractions}, *=modified main verb.}
	\label{tab:modality_ie_mapping}
\end{table}

\end{document}